\documentclass[twocolumn,english,letterpaper,10pt,journal,twoside]{IEEEtran}
\usepackage[T1]{fontenc}
\usepackage[latin9]{inputenc}
\usepackage{array}
\usepackage{prettyref}
\usepackage{refstyle}
\usepackage{float}
\usepackage{multirow}
\usepackage{amsmath}
\usepackage{amsthm}
\usepackage{amssymb}
\usepackage{stmaryrd}
\usepackage{graphicx}

\makeatletter

\AtBeginDocument{\providecommand\defref[1]{\ref{def:#1}}}
\AtBeginDocument{\providecommand\eqref[1]{\ref{eq:#1}}}
\AtBeginDocument{\providecommand\lemref[1]{\ref{lem:#1}}}
\AtBeginDocument{\providecommand\thmref[1]{\ref{thm:#1}}}

\providecommand{\tabularnewline}{\\}
\RS@ifundefined{subsecref}
  {\newref{subsec}{name = \RSsectxt}}
  {}
\RS@ifundefined{thmref}
  {\def\RSthmtxt{theorem~}\newref{thm}{name = \RSthmtxt}}
  {}
\RS@ifundefined{lemref}
  {\def\RSlemtxt{lemma~}\newref{lem}{name = \RSlemtxt}}
  {}

\theoremstyle{plain}
\newtheorem{thm}{\protect\theoremname}
\theoremstyle{definition}
\newtheorem{example}[thm]{\protect\examplename}
\theoremstyle{definition}
\newtheorem{defn}[thm]{\protect\definitionname}
\theoremstyle{plain}
\newtheorem{lem}[thm]{\protect\lemmaname}
\newenvironment{lyxcode}
{\par\begin{list}{}{
\setlength{\rightmargin}{\leftmargin}
\setlength{\listparindent}{0pt}
\raggedright
\setlength{\itemsep}{0pt}
\setlength{\parsep}{0pt}
\normalfont\ttfamily}
 \item[]}
{\end{list}}

\IEEEoverridecommandlockouts

\usepackage[tracking=false,kerning=true,spacing=true]{microtype}

\usepackage[table]{xcolor}

\usepackage{wrapfig}
\usepackage{prettyref}
\newrefformat{prob}{Problem~\ref{#1}}
\newrefformat{sub}{Section~\ref{#1}}
\newrefformat{prop}{Proposition~\ref{#1}}
\newrefformat{def}{Definition~\ref{#1}}
\newrefformat{app}{Appendix~\ref{#1}}
\newrefformat{alg}{Algorithm~\ref{#1}}
\newrefformat{cor}{Corollary~\ref{#1}}
\newrefformat{thm}{Theorem~\ref{#1}}
\newrefformat{tab}{Table~\ref{#1}}
\newrefformat{fig}{Fig.~\ref{#1}}
\newrefformat{sec}{Section~\ref{#1}}
\newrefformat{def}{Def.~\ref{#1}}
\newrefformat{line}{line~\ref{#1}}
\newrefformat{code}{Listing~\ref{#1}}

\newcommand*{\minwidthbox}[2]{
  \makebox[{\ifdim#2<\width\width\else#2\fi}]{#1}
}

\newcommand{\tableColors}{\rowcolors{2}{green!4}{blue!4}}

\let\l@ENGLISH\l@english

\usepackage[style=numeric-comp,sorting=none,firstinits=true, maxnames=10, bibstyle=numeric,abbreviate=true,defernums=true,eprint=false,backend=bibtex]{biblatex}
\providecommand{\weburl}[1]{}
\providecommand{\biburl}[1]{}

\newcommand{\notpresent}[1]{{\color{red}No #1} }
\renewcommand{\notpresent}[1]{}

\bibliography{commons/bib/strings_long}
\bibliography{commons/bib/all_only_one_w_pages}

\renewcommand{\weburl}[1]{\href{#1}{(url)}}
\renewcommand{\biburl}[1]{\textrm{\href{#1}{#1}}}

\RequirePackage{xcolor}
\definecolor{darkred}{rgb}{0.3,0,0}
\definecolor{darkgreen}{rgb}{0,0.3,0}
\definecolor{urlcolor}{rgb}{0,0,0.6}

\RequirePackage{hyperref}
\hypersetup{bookmarks=true,colorlinks=true,urlcolor=urlcolor,citecolor=darkgreen,linkcolor=darkred}

\usepackage{xspace}

\usepackage{amsmath}
\usepackage{amssymb}
\usepackage[mathscr]{eucal}

\usepackage[caption=false,font=footnotesize]{subfig}
\usepackage{graphicx}

\newcommand{\reals}{\mathbb{R}}

\newcommand{\aword}[1]{\mathsf{#1}}

\newcommand{\vmath}[1]{\aword{#1}}

\DeclareMathOperator*{\Min}{Min}

\newcommand{\posleq}{\preceq}

\newcommand{\poslt}{\prec}

\newcommand{\posA}{\mathcal{P}}

\newcommand{\lfp}{\vmath{lfp}}

\newcommand{\antichains}{\vmath{A}}

\newcommand{\upsets}{\vmath{U}}

\newcommand{\ftor}{{h}}

\usepackage[normalem]{ulem}

\newcommand{\funsp}{\mathscr{F}} 
 
\newcommand{\fun}{\vmath{f}}

\newcommand{\res}{\vmath{r}}

\newcommand{\ressp}{\mathscr{R}}

\newcommand{\dprob}{\vmath{dp}}

\newcommand{\dpseries}{\vmath{series}}
\newcommand{\dppar}{\vmath{par}}
\newcommand{\dploop}{\vmath{loop}}

\newcommand{\Aressp}{{\antichains\ressp}}

\newcommand{\udpa}{\boldsymbol{u}_a}
\newcommand{\udpb}{\boldsymbol{u}_b}
\newcommand{\udpL}{\boldsymbol{\mathsf{L}}}
\newcommand{\udpU}{\boldsymbol{\mathsf{U}}}
\newcommand{\udpsp}{\vmath{UDP}}
\newcommand{\udpleq}{\posleq_\udpsp}

\newcommand{\dpsp}{\vmath{DP}}
\newcommand{\dpleq}{\posleq_\dpsp}

\newcommand{\terms}{\vmath{Terms}}

\newcommand{\udpsem}{\Phi}

\newcommand{\dpsem}{\varphi}

\newcommand{\atoms}{\mathcal{A}}
\newcommand{\atree}{\boldsymbol{\vmath{T}}}
\newcommand{\val}{\boldsymbol{v}} 

\newcommand{\ops}{\vmath{ops}}

\newcommand{\acprod}{\mathbin{\boldsymbol{\times}}} 

\newcommand{\oploop}{\dagger}
\newcommand{\opseries}{\mathbin{\varocircle}}
\newcommand{\oppar}{\mathbin{\varotimes}}

\newcommand{\UId}{\vmath{UId}} 
\newcommand{\vdc}{\vmath{vdc}}

\newcommand{\colR}{\color[rgb]{0.555789,0.000000,0.000000}}
\newcommand{\colF}{\color[rgb]{0.094869,0.500000,0.000000}}
\newcommand{\colH}{\color[rgb]{0.000000,0.400000,1.000000}}

\newcommand{\colU}{\color{purple}}
\newcommand{\colL}{\color{orange}}

\newcommand{\R}[1]{{\colR #1}}
\newcommand{\F}[1]{{\colF #1}}

\usepackage{wrapfig}

\usepackage{stmaryrd}

\newcommand{\person}[2]{#1\,{\footnotesize<#2>\rm}}

\newcommand{\mythanks}{

\thanks{This paper was recommended for publication by Editor Kevin Lynch upon evaluation of the Associate Editor and Reviewers' comments.}
\thanks{This work was supported by the National Science Foundation,
National Robotics Initiative, award 1405259.}
  \protect\thanks{\person{Andrea\,Censi}{acensi@ethz.ch} is with the
  Institute for Dynamic Systems and Control at ETH Z\"urich.
  This work was performed while the author was with the
  Laboratory for Information and Decision Systems (LIDS)
  at the Massachusetts Institute of Technology.
}
\thanks{Digital Object Identifier (DOI): see top of this page.}
}

\usepackage{capt-of}

\usepackage{adjustbox}

\usepackage{pdfpages}

\newcommand{\C}[1]{#1}

\usepackage{enumitem}

\setlist[itemize]{leftmargin=0pt,itemindent=1em}
\setlist[enumerate]{leftmargin=3pt,itemindent=1em}

\let\oldfun\fun     \renewcommand{\fun}{{\colF\oldfun}}
\let\oldres\res     \renewcommand{\res}{{\colR\oldres}}

\let\oldfunsp\funsp \renewcommand{\funsp}{{\colF\oldfunsp}}
\let\oldressp\ressp \renewcommand{\ressp}{{\colR\oldressp}}

\let\oldftor\ftor \renewcommand{\ftor}{{\colH\oldftor}}

\renewcommand{\Aressp}{{\colR\antichains\ressp}}

\let\oldudpL\udpL \renewcommand{\udpL}{{\colL\oldudpL}}
\let\oldudpU\udpU \renewcommand{\udpU}{{\colU\oldudpU}}

\newcommand{\ufloor}{{\colL\vmath{floor}}}
\newcommand{\uceil}{{\colU\vmath{ceil}}}

\@ifundefined{showcaptionsetup}{}{
 \PassOptionsToPackage{caption=false}{subfig}}
\usepackage{subfig}
\makeatother

\usepackage{babel}
\providecommand{\definitionname}{Definition}
\providecommand{\examplename}{Example}
\providecommand{\lemmaname}{Lemma}
\providecommand{\theoremname}{Theorem}

\begin{document}

\title{Uncertainty in Monotone Co-Design Problems}

\author{Andrea Censi\mythanks}
\maketitle
\begin{abstract}
This work contributes to a compositional theory of ``co-design''
that allows to optimally design a robotic platform. In this framework,
the user describes each subsystem as a monotone relation between \F{``functionality''
provided} and \R{``resources'' required}. These models can be
easily composed to express the co-design constraints among different
subsystems. The user then queries the model, to obtain the design
with minimal resources usage, subject to a lower bound on the provided
functionality. This paper concerns the introduction of uncertainty
in the framework. Uncertainty has two roles: first, it allows to deal
with limited knowledge of the models; second, it also can be used
to generate consistent relaxations of a problem, as the computation
requirements can be lowered, should the user accept some uncertainty
in the answer.
\end{abstract}


\section{Introduction}

\begin{figure}[!t]
\begin{centering}
\subfloat[\label{fig:Example1}Part of a co-design diagram for a UAV.]{\begin{centering}
\includegraphics[width=1\columnwidth]{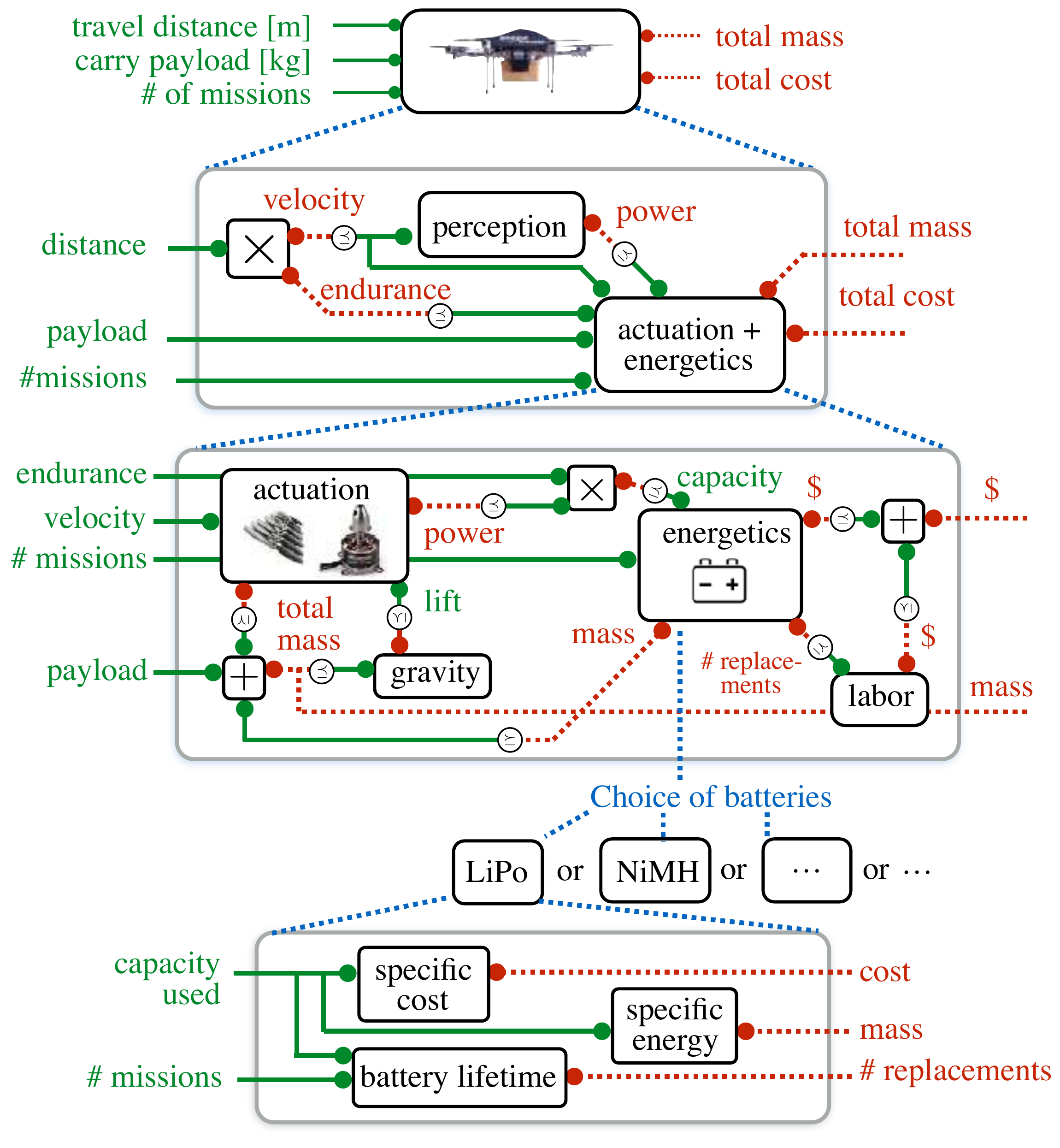}
\par\end{centering}
}
\par\end{centering}
\begin{centering}
\subfloat[\label{fig:udp-bounds}Uncertain design problems]{\begin{centering}
\hspace{1cm}\includegraphics[scale=0.33]{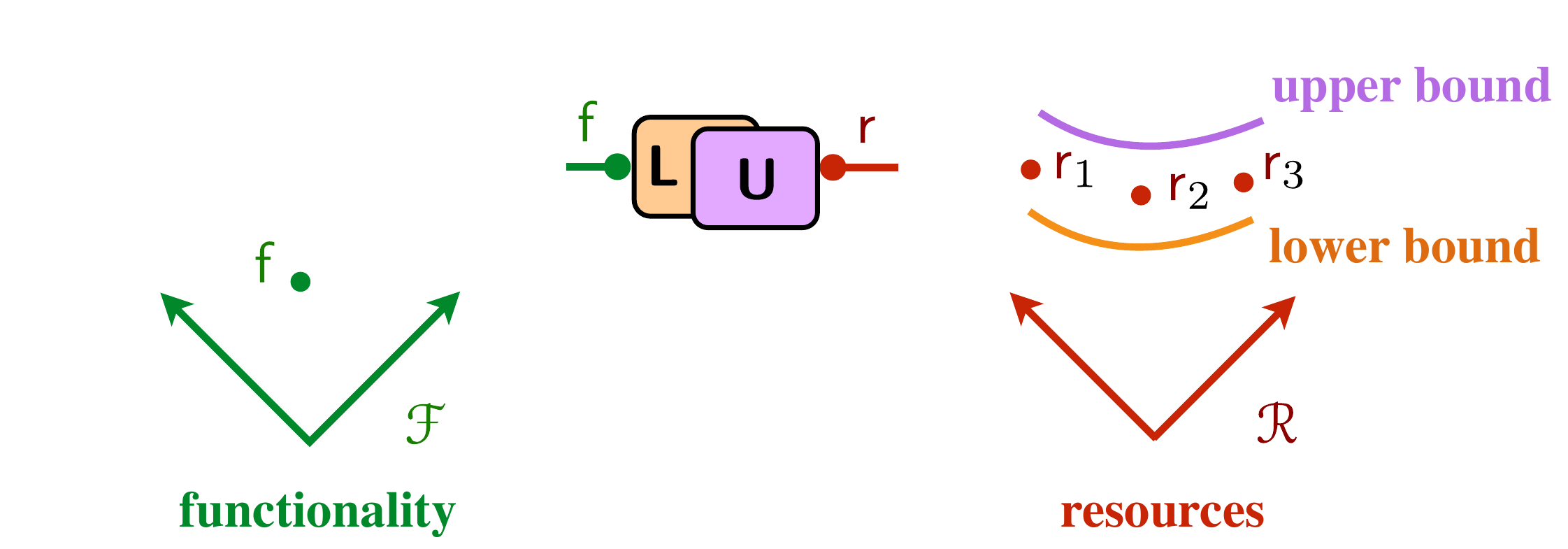}
\par\end{centering}
}
\par\end{centering}
\caption{\emph{Panel} \emph{a}: Monotone Co-Design Problems capture the co-design
constraints among the components of a complex design, by describing
\emph{design problems }for each component in isolation and\emph{ co-design
constraints }among different components. The co-design diagram in
the figure describes part of the design problem for a UAV, and in
particular how the \emph{resources} (\R{total mass} and \R{cost})
are related to the \emph{functionality} (\F{distance}, \F{payload},
\F{number of missions}). Functionality edges are green and solid;
resources are red and dotted. \emph{Panel} \emph{b}:~This paper describes
how to introduce uncertainty in this framework, which allows, for
example, to introduce parametric uncertainty in the definition of
components properties (e.g. specific cost of batteries). An Uncertain
Design Problem (UDP) is described by a pair of functions from functionality
to subset of resources that give an upper and lower bound on the resource
consumption.}
\end{figure}

\IEEEPARstart{T}{he} design of a robotic platform involves the choice
and configuration of many hardware and software subsystems (actuation,
energetics, perception, control, \ldots{}) in an harmonious entity.
Because robotics is a young discipline, there is still little work
towards obtaining systematic procedures to derive optimal designs.
Therefore, robot design is a lengthly process based on empirical evaluation
and trial and error. 

\C{The work presented here contributes to a theory of co-design~\cite{censi15same,censi16codesign_sep16}
that allows to optimally design a robotic platform based on formal
models of its subsystems. The goal is to allow a designer to create
better designs, faster. This work on ``co-design'' is related to
and complementary to works that deal with ``co-generation'' (the
ability of synthesizing hardware and software blueprints for entire
robot platforms)~\cite{mehta14cogeneration,mehta15robot}. }

\C{This paper describes the introduction of uncertainty in the theory
developed so far. In this framework, the user defines ``design problems''
for each physical or logical subsystem. Each design problem (DP) is
a relation between ``\F{functionality}'' provided and ``\R{resources}''
required by the component. The DPs can then composed in a graph, where
each edge represents a ``co-design constraint'' between two DPs\emph{.}
The resulting class of problems is called Monotone Co-Design Problems
(MCDPs)}.

An example of MCDP is sketched in~\prettyref{fig:Example1}. The
design problem consists in finding an optimal configuration of a UAV,
optimizing over actuators, sensors, processors, and batteries. \C{In
this simplified example, the functionality of the UAV is parameterized
by three numbers}: the \F{distance to travel} for each mission;
the \F{payload to transport}; the \F{number of missions} to fly.
The optimal design is defined as the one that satisfies the functionality
constraints while using the minimal amount of \R{resources} (\R{cost}
and \R{mass}). In the figure, the model is exploded to show how
actuation and energetics are modeled. Perception is modeled as a relation
between \F{the velocity of the platform} and the \R{power} required
\C{(the faster the platform, the more data needs to be processed)}.
Actuation is modeled as a relation between \F{lift} and \R{power}/\R{cost}.
Batteries are described by a relation between \F{capacity} and \R{mass}/\R{cost}.
In this example, there are different battery technologies (LiPo, etc.),
each specified by specific energy, specific cost, and lifetime, thus
characterized by a different relation between \F{capacity}, \F{number
of missions} and \R{mass} and \R{cost}. The interconnection between
design problems describe the ``co-design constraints'', which could
be recursive: e.g., actuators must lift the batteries, the batteries
must power the actuators. Cycles represent design problems that are
coupled.

Once the model is defined, it can be queried to obtain the \emph{minimal}
solution in terms of resources \textemdash{} here, \R{total cost}
and \R{total mass}. The output to the user is the Pareto front containing
all non-dominated solutions. The corresponding optimization problem
is, in general, nonconvex. Yet, with few assumptions, it is possible
to obtain a systematic solution procedure, and show that there exists
a dynamical system whose \C{fixed point corresponds} to the set
of minimal solutions. 

This paper describes how to add a notion of \emph{uncertainty} in
the MCDP framework. The model of uncertainty considered is interval
uncertainty on arbitrary partial orders. For a partially ordered set
(poset)~$\left\langle \posA,\posleq\right\rangle $, these are sets
of the type~$\{x\in\posA\colon a\posleq x\posleq b\}$. I will show
how one can introduce this type of uncertainty in the MCDP framework
by considering ordered pairs of design problems. Each pair describes
lower and upper bounds for resources usage. These \emph{uncertain
design problems} (UDPs) can be composed using series, parallel, and
feedback interconnection, just like their non-uncertain counterparts. 

\C{The output to the user is \emph{two} Pareto fronts, describing
the minimal resource consumptions in the best case and in the worst
case according to the models specified. One or both the Pareto fronts
can be empty, meaning that the problem does not have a feasible solution.} 

This is different from the usual formalization of ``robust optimization''~\cite{bertsimas11theory,ben-tal09},
usually formulated as a ``worst case'' analysis, in which the uncertainty
in the problem is described by a set of possible parameters, and the
optimization problem is posed as finding the one design that is valid
for all cases. 

Uncertainty plays two roles: it can be used as a \emph{modeling} \emph{tool},
where the relations are uncertain because of our limited knowledge,
and it can be used as a \emph{computational} \emph{tool}, in which
we deliberately choose to consider uncertain relations as a relaxation
of the problem, to reduce the computational load, while maintaining
precise consistency guarantees. With these additions, the MCDP framework
\C{can} describe even richer design problems and to efficiently
solve them.

\subsubsection*{Paper organization}

Section \ref{sec:Design-Problems} and \ref{sec:Monotone-Co-Design-Problems}
summarize previous work. They give a formal definition of design problems
(DPs) and their composition, called Monotone Co-Design Problems (MCDPs).
Section~\ref{sec:UDP} through~\ref{sec:Approximation-results}
describe the notion of Uncertain Design Problem (UDP), the semantics
of their interconnection, and the general theoretical results. Section~\ref{sec:Applications}
describes three specific applications of the theory with numerical
results. The supplementary materials (also available as \cite{mcdp_icra_uncertainty_extended})
include detailed models written in MCDPL and pointers to obtain the
source code and a virtual machine for reproducing the experiments.

\section{Design Problems\label{sec:Design-Problems}}

\emph{A design problem} (DP) is a monotone relation between \emph{\F{provided functionality}}
and \emph{\R{required resources}}. \F{Functionality} and \R{resources}
are complete partial orders (CPO)~\cite{davey02}, denoted~$\langle\funsp,\posleq_{\funsp}\rangle$
and~$\langle\ressp,\posleq_{\ressp}\rangle$. The graphical representations
uses nodes for DPs and green and red edges for \F{functionality}
and \R{resources}.

\begin{figure}[H]
\begin{centering}
\includegraphics[scale=0.33]{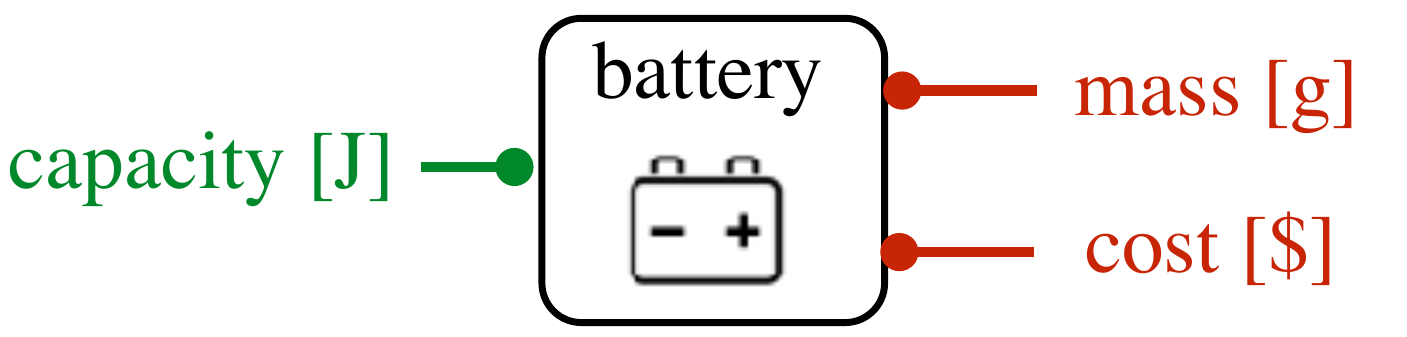}
\par\end{centering}
\caption{\label{fig:battery1}Modeling a battery as a \emph{design problem.}}
\end{figure}

\begin{example}
The first-order characterization of a battery is as a store of energy,
in which the \F{capacity {[}kWh{]}} is the \F{functionality} (what
the battery provides) and \R{mass} {[}kg{]} and \R{cost} {[}\${]}
are \R{resources} (what the battery requires)~(\prettyref{fig:battery1}). 
\end{example}
\noindent For a fixed functionality~$\fun\in\funsp$, the set of
minimal resources in~$\ressp$ sufficient to perform the functionality
might contain two or more elements that are incomparable with respect
to~$\posleq_{\ressp}$. For example, in the case of a battery, one
might consider different battery technologies that are incomparable
in the \R{mass}/\R{cost} resource space. 

A subset with ``minimal'', ``incomparable'' elements is called
``antichain''. This is the mathematical formalization of what is
informally called a ``Pareto front''.

\begin{figure}[H]
\begin{centering}
\includegraphics[bb=0bp 0bp 160bp 72bp,clip,scale=0.33]{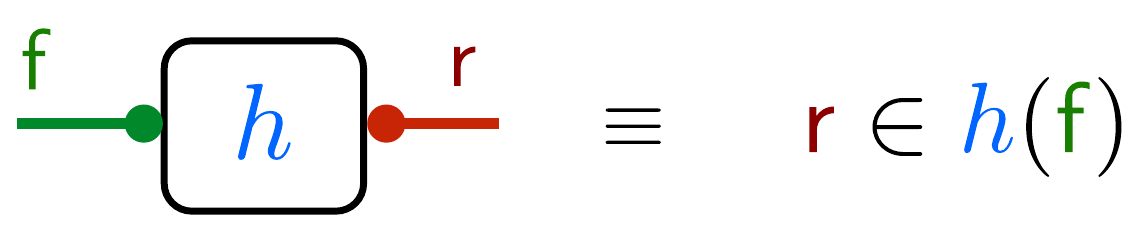}
\par\end{centering}
\begin{centering}
\includegraphics[scale=0.33]{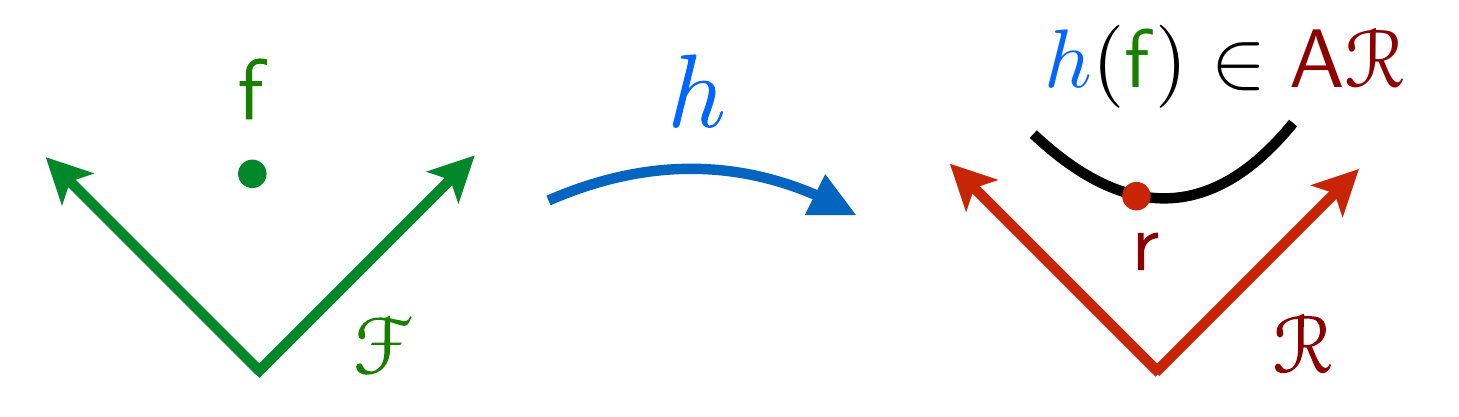}
\par\end{centering}
\caption{\C{\label{fig:antichain}A DP is a represented by a map ${\colH\ftor}:{\colF\funsp}\rightarrow{\colR\antichains\ressp}$
from functionality to antichains of resources.}}
\end{figure}

\begin{defn}
An \emph{antichain}~$S$ in a poset~$\left\langle \posA,\posleq\right\rangle $
is a subset of~$\posA$ such that no element of~$S$ dominates another
element: \C{for all}~$x,y\in S$ and~$x\posleq y$, then~$x=y$. 
\end{defn}
\begin{lem}
Let~$\antichains\posA$ be the set of antichains of~$\posA$. $\antichains\posA$
is a poset itself, with the partial order~$\posleq_{\antichains\posA}$
defined as
\begin{equation}
S_{1}\posleq_{\antichains\posA}S_{2}\ \equiv\ \uparrow S_{1}\supseteq\,\uparrow S_{2},\label{eq:orderantichains}
\end{equation}
where ``$\uparrow$'' denotes the upper closure of a set.
\end{lem}
\begin{defn}
\label{def:A-monotone-design}A\emph{ design problem~}(DP) is a tuple~$\left\langle \funsp,\ressp,\ftor\right\rangle $
such that~$\funsp$ and~$\ressp$ are CPOs, and~${\colH\ftor}:{\colF\funsp}\rightarrow{\colR\antichains\ressp}$
is a monotone and Scott-continuous function~(\cite{gierz03continuous}
or \cite[Definition 11]{censi16codesign_sep16}). Each functionality~$\fun$
\C{(or vector of functionalities, if $\funsp$ is a product of posets)}
corresponds to an antichain of resources~$\ftor(\fun)\in\Aressp$~(\prettyref{fig:antichain}). 
\end{defn}

\noindent Monotonicity implies that, if the functionality is increased,
then the required resources increase as well.

\noindent

\section{Monotone Co-Design Problems \label{sec:Monotone-Co-Design-Problems}}

A Monotone Co-Design Problem (MCDP) is a multigraph of DPs. \C{An
edge between a resource~$\res_{1}$ of a DP and a functionality~$\fun_{2}$
of another denotes the partial order inequality constraint~$\res_{1}\posleq\fun_{2}$.
Cycles and self-loops are allowed.}

\begin{example}
The MCDP in~\prettyref{fig:example} is the interconnection of 3
DPs~$\ftor_{a},\ftor_{b},\ftor_{c}.$ The semantics as an optimization
problem is shown in~\prettyref{fig:example-semantics}. We will also
use an ``algebraic'' representation, shown in \prettyref{fig:example-b},
and defined in \prettyref{def:MCDP-algebraic}.\\
\begin{figure}[H]
\begin{centering}
\subfloat[\label{fig:example}Graphical representation]{\begin{centering}
\includegraphics[scale=0.37]{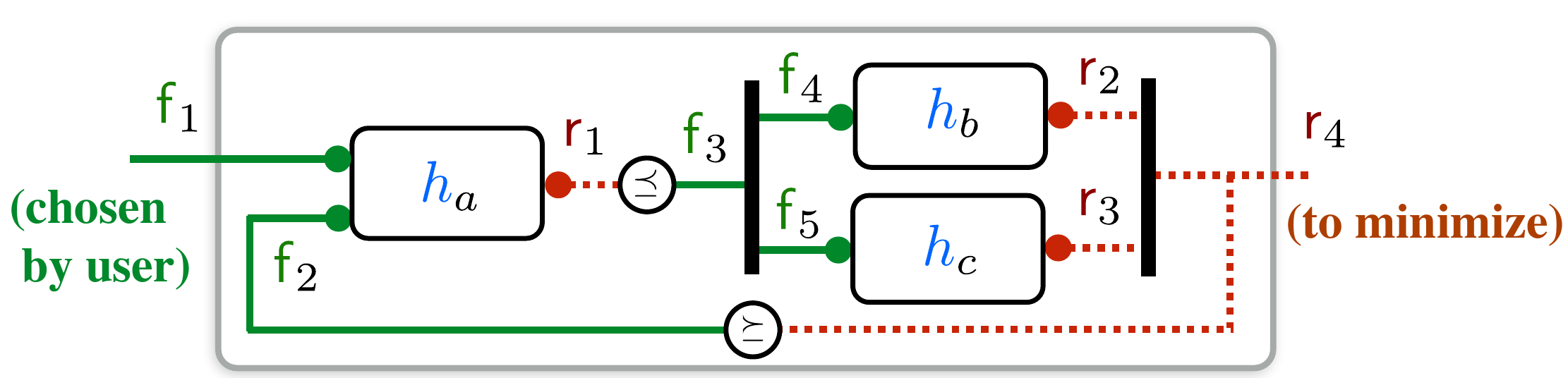}
\par\end{centering}
}\\
\subfloat[\label{fig:example-semantics}Semantics as an optimization problem]{\begin{centering}
\includegraphics[scale=0.4]{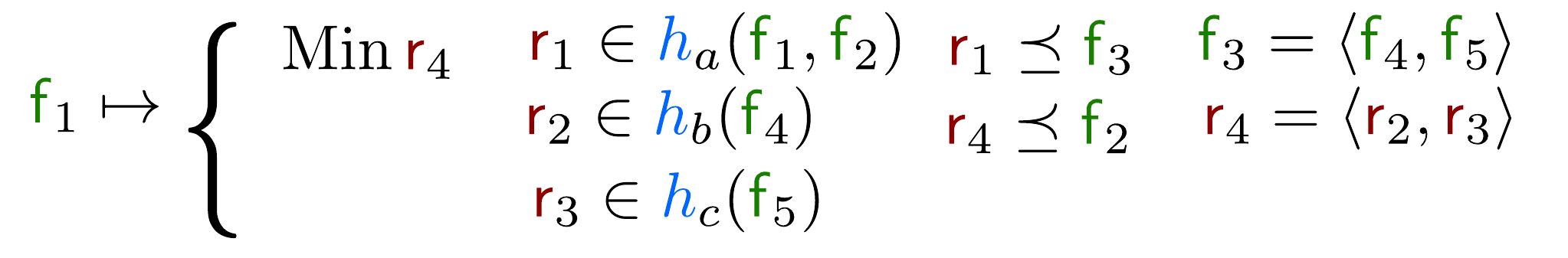}
\par\end{centering}
}\\
\subfloat[\label{fig:example-b}Algebraic representation]{\begin{centering}
\includegraphics[scale=0.33]{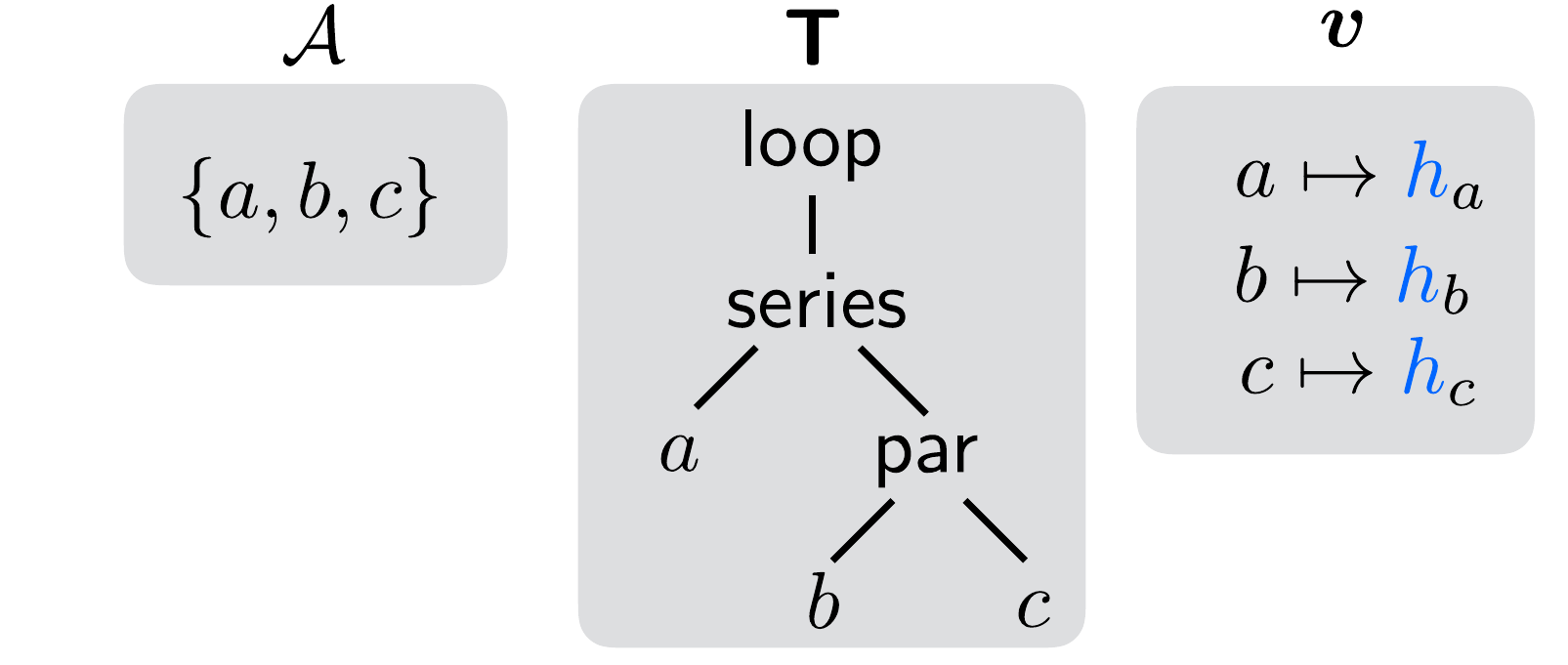}
\par\end{centering}
}
\par\end{centering}
\caption{Three equivalent representations of an MCDP.}
\end{figure}
\end{example}
\C{The functionality/resources parametrization is quite natural for
many design engineering domains. Moreover, it allows for quantitative
optimization, in contrast to qualitative modeling tools such as \emph{function
structure }diagrams~\cite{pahl07}}. All models considered may
be nonlinear, in contrast to work such as Suh's theory of \emph{axiomatic
design~}\cite{suh01}.

\subsection{Algebraic definition}

\begin{figure}[H]
\centering{}\hfill{}\subfloat[$\dpseries(a,b)$]{\includegraphics[scale=0.33]{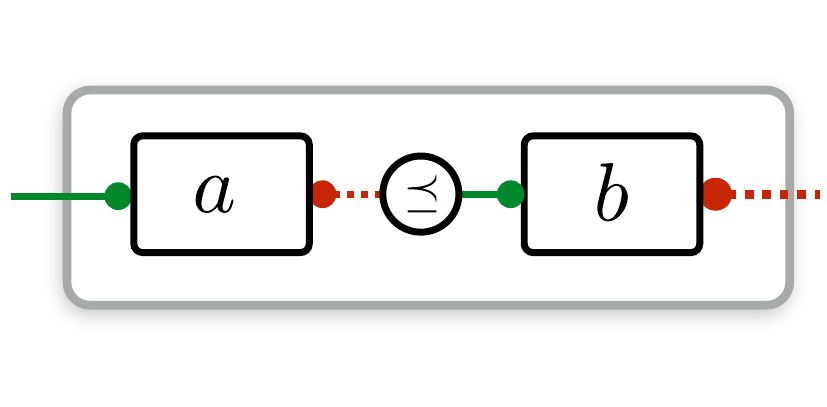}

}\hfill{}\subfloat[$\dppar(a,b)$]{\includegraphics[scale=0.33]{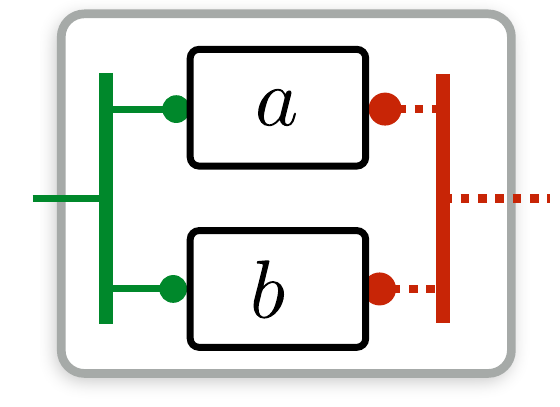}

}\hfill{}\subfloat[$\dploop(a)$]{\includegraphics[scale=0.33]{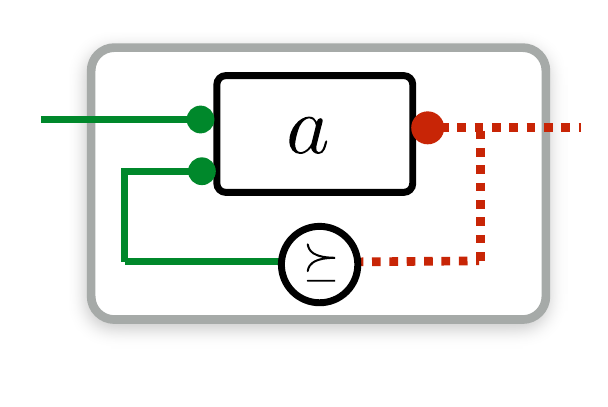}

}\smallskip{}
\caption{\label{fig:series-par-loop}The three operators used in the inductive
definition of MCDPs.}
\end{figure}

Some of the proofs rely on an algebraic representation of the graph.
Series-parallel graphs (see, e.g.,~\cite{duffin65topology}) have
widespread use in computer science. Here, we add a third operator
to be able to represent loops. In the algebraic definition, the graph
is a represented by a tree, where the leaves are the nodes, and the
junctions are one of three operators ($\dpseries,\dppar,\dploop$),
as in~\prettyref{fig:series-par-loop}. An equivalent construction
for network processes is given in Stefanescu~\cite{stefanescu00}.
Equivalently, we are defining a symmetric traced monoidal category
(see, e.g.,~\cite{joyal96traced} or~\cite{spivak14category} for
an introduction); note that the~$\dploop$ operator is related to
the ``trace'' operator, but not exactly equivalent, though they
can be defined in terms of each other. 

Let us use a standard definition of ``operators'', ``terms'',
and ``atoms'' (see, e.g., ~\cite[p.41]{jezek08}). Given a set
of operators~$\ops$ and a set of atoms~$\atoms$, let~$\terms(\ops,\atoms)$
be the set of all inductively defined expressions. \C{For example,
if the operator set contains only one operator~$f$ with one argument,
and there is only one atom~$a$,} then the terms are $\terms(\{f\},\{a\})=\{a,f(a),f(f(a)),\dots\}.$

\begin{defn}[Algebraic definition of Monotone Co-Design Problems]
\label{def:MCDP-algebraic}An MCDP is a tuple~$\left\langle \atoms,\atree,\val\right\rangle $,
where:
\begin{enumerate}
\item $\atoms$ is any set of atoms, to be used as labels.
\item The term~$\atree$ in the $\{\dpseries,\dppar,\dploop\}$ algebra
describes the structure of the graph:
\[
\atree\in\terms(\{\dpseries,\dppar,\dploop\},\atoms).
\]
\item The \emph{valuation} $\val:\atoms\rightarrow\dpsp$ assigns a DP to
each atom.
\end{enumerate}
\end{defn}
\begin{example}
The MCDP in~\prettyref{fig:example} can be described by the atoms
$\atoms=\{a,b,c\}$, the term $\atree=\dploop(\dpseries(a,\dppar(b,c)),$
plus the valuation $\val:\{a\mapsto\ftor_{a},b\mapsto\ftor_{b},c\mapsto\ftor_{c}\}.$
The tuple~$\left\langle \atoms,\atree,\val\right\rangle $ for this
example is shown in \prettyref{fig:example-b}.
\end{example}

\subsection{Semantics of MCDPs}

We can now define the \emph{semantics} of an MCDP. The \emph{semantics}
is a function~$\dpsem$ that, given an algebraic definition of an
MCDP, returns a~$\dpsp$. Thanks to the algebraic definition, to
define~$\dpsem$, we need to only define what happens in the base
case (equation~\ref{eq:base}), and what happens for each operator
$\dpseries,\dppar,\dploop$ (equations~\ref{eq:series}\textendash \ref{eq:loop}).
\begin{defn}[Semantics of MCDP]
\label{def:dpsem}Given an MCDP in algebraic form~$\left\langle \atoms,\atree,\val\right\rangle $,
the semantics
\[
\dpsem\llbracket\left\langle \atoms,\atree,\val\right\rangle \rrbracket\in\dpsp
\]
is defined as follows: \\
\adjustbox{max width=8.6cm}{{\small{}}
\begin{minipage}[t]{1.05\columnwidth}
{\small{}
\begin{align}
\dpsem\llbracket\left\langle \atoms,a,\val\right\rangle \rrbracket & \doteq\val(a),\qquad\text{for all}\ a\in\atoms,\label{eq:base}\\
\dpsem\llbracket\left\langle \atoms,\dpseries(\atree_{1},\atree_{2}),\val\right\rangle \rrbracket & \doteq\dpsem\llbracket\left\langle \atoms,\atree_{1},\val\right\rangle \rrbracket\,\opseries\,\dpsem\llbracket\left\langle \atoms,\atree_{2},\val\right\rangle \rrbracket,\label{eq:series}\\
\dpsem\llbracket\left\langle \atoms,\dppar(\atree_{1},\atree_{2}),\val\right\rangle \rrbracket & \doteq\dpsem\llbracket\left\langle \atoms,\atree_{1},\val\right\rangle \rrbracket\,\oppar\,\dpsem\llbracket\left\langle \atoms,\atree_{2},\val\right\rangle \rrbracket,\label{eq:par}\\
\dpsem\llbracket\left\langle \atoms,\dploop(\atree),\val\right\rangle \rrbracket & \doteq\dpsem\llbracket\left\langle \atoms,\atree,\val\right\rangle \rrbracket^{\oploop}.\label{eq:loop}
\end{align}
}
\end{minipage}{\small{}}}{\small \par}
\end{defn}
The operators $\opseries,\oppar,\oploop$ are defined \C{in~\prettyref{def:opmaps}\textendash \prettyref{def:oploop}}.
Please see~\cite[Section VI]{censi16codesign_sep16} for details
about the interpretation of these operators and how they are derived. 

The $\oppar$ operator is a regular product in category theory: we
are considering all possible combinations of resources required by~$\ftor_{1}$
and~$\ftor_{2}$.
\begin{defn}[Product operator $\oppar$]
\label{def:opmaps}For two maps $\ftor_{1}\colon\funsp_{1}\rightarrow\Aressp_{1}$
and $\ftor_{2}\colon\funsp_{2}\rightarrow\Aressp_{2}$, define
\begin{align*}
\ftor_{1}\oppar\ftor_{2}:(\funsp_{1}\times\funsp_{2}) & \rightarrow\antichains(\ressp_{1}\times\ressp_{2}),\\
\left\langle \fun_{1},\fun_{2}\right\rangle  & \mapsto\ftor_{1}(\fun_{1})\acprod\ftor_{2}(\fun_{2}),
\end{align*}
where $\acprod$ is the product of two antichains.
\end{defn}
The $\opseries$ operator is similar to a convolution: fixed $\fun_{1}$,
one evaluates the resources $\res_{1}\in\ftor_{1}(\fun)$, and for
each~$\res_{1}$, $\ftor_{2}(\res_{1})$ is evaluated. Then the minimal
elements are selected.
\begin{defn}[Series operator~$\opseries$]
\label{def:opseries}For two maps~$\ftor_{1}\colon\funsp_{1}\rightarrow\Aressp_{1}$
and~$\ftor_{2}\colon\funsp_{2}\rightarrow\Aressp_{2}$, if~$\ressp_{1}=\funsp_{2}$
, define
\begin{align*}
{\displaystyle \ftor_{1}\opseries\ftor_{2}\colon\funsp_{1}} & \rightarrow\Aressp_{2},\\
\ftor_{1} & \mapsto\Min_{\posleq_{\ressp_{2}}}\bigcup_{\res_{1}\in\ftor_{1}(\fun)}\ftor_{2}(\res_{1}).
\end{align*}

\end{defn}

The dagger operator $\oploop$ is actually a standard operator used
in domain theory (see, e.g., \cite[II-2.29]{gierz03continuous}).
\begin{defn}[Loop operator $\oploop$]
\label{def:oploop}For a map $\ftor:\funsp_{1}\times\funsp_{2}\rightarrow\Aressp$,
define
\begin{align}
\ftor^{\oploop}:\funsp_{1} & \rightarrow\Aressp,\nonumber \\
\fun_{1} & \mapsto\lfp\left(\Psi_{\fun_{1}}^{\ftor}\right),\label{eq:lfp}
\end{align}
where $\lfp$ is the least-fixed point operator, and~$\Psi_{\fun_{1}}^{\ftor}$
is
\begin{align*}
\Psi_{\fun_{1}}^{\ftor}:\Aressp & \rightarrow\Aressp,\\
{\colR R} & \mapsto\Min_{\posleq_{\ressp}}\bigcup_{\res\in{\colR R}}\ftor(\fun_{1},\res)\ \cap\uparrow\res.
\end{align*}
\end{defn}

\subsection{Solution of MCDPs}

\prettyref{def:dpsem} gives a way to evaluate the map~$\ftor$ for
the graph, given the maps~$\{\ftor{}_{a}\mid a\in\atoms\}$ for the
leaves. Following those instructions, we can compute~$\ftor(\fun)$,
and thus find the minimal resources needed for the entire MCDP. 
\begin{example}
The MCDP in~\prettyref{fig:example} is so small that we can do this
explicitly. From~\prettyref{def:dpsem}, we can compute the semantics
as follows:
\begin{align*}
\ftor & =\dpsem\left\llbracket \langle\atoms,\dploop(\dpseries(a,\dppar(b,c)),\val\rangle\right\rrbracket \\
 & =\left(\ftor_{a}\,\opseries\,\left(\ftor_{b}\,\oppar\,\ftor_{c}\right)\right)^{\oploop}.
\end{align*}
Substituting the definitions~\ref{def:opmaps}\textendash \ref{def:oploop}
above, one finds that $\ftor(\fun)=\lfp\left(\Psi_{\fun}\right),$
with
\begin{align*}
\Psi_{\fun}:\Aressp & \rightarrow\Aressp,\\
{\colR R} & \mapsto\bigcup_{\res\in{\colR R}}\Big[\Min_{\posleq}\uparrow\bigcup_{s\in\ftor_{a}(\fun_{1},\res)}\ftor_{b}(s)\acprod\ftor_{c}(s)\Big]\ \cap\uparrow\res.
\end{align*}
The least fixed point equation can be solved using Kleene's algorithm~\cite[CPO Fixpoint theorem I, 8.15]{davey02}.
A dynamical system that computes the set of solutions is given by
\[
\begin{cases}
{\colR R}_{0} & \leftarrow\{\bot_{\ressp}\},\\
{\colR R}_{k+1} & \leftarrow\Psi_{\fun}({\colR R}_{k}).
\end{cases}
\]
The limit $\sup{\colR R}_{k}$ is the set of minimal solutions, which
might be an empty set if the problem is unfeasible for a particular
value~$\fun$.

This dynamical system is a proper algorithm only if each step can
be performed with bounded computation. An example in which this is
not the case are relations that give an infinite number of solutions
for each functionality. For example, the very first DP appearing in~\prettyref{fig:Example1}
corresponds to the relation~\emph{${\colF\text{travel distance}}\leq{\colR\text{velocity}}\times{\colR\text{endurance}},$}
for which there are infinite numbers of pairs~$\langle{\colR\text{velocity}},{\colR\text{endurance}}\rangle$
for each value of~${\colF\text{travel distance}}$. The machinery
developed in this paper will make it possible to deal with these infinite-cardinality
relations by relaxation.
\end{example}

\section{Uncertain Design Problems\label{sec:UDP}}

This section describes the notion of Uncertain DPs (UDPs). UDPs are
an ordered pair of DPs that can be interpreted as upper and lower
bounds for resource consumptions~(\prettyref{fig:udp-bounds}). We
will be able to propagate this interval uncertainty through an arbitrary
interconnection of DPs. The result presented to the user will be a
\emph{pair} of antichains \textemdash{} a lower and an upper bound
for the resource consumption. 

\subsection{Partial order $\dpleq$}

Being able to provide both upper and lower bounds comes from the fact
that in this framework everything is ordered \textendash{} there is
a poset of resources, lifted to posets of antichains, which is lifted
to posets of DPs, and finally, to the poset of uncertain DPs. 

The first step is defining a partial order~$\dpleq$ on~$\dpsp$.
\begin{defn}[Partial order $\dpleq$]
Consider two DPs $\ftor_{1},\ftor_{2}:\funsp\rightarrow\Aressp$.
The DP~$\ftor_{1}$ precedes~$\ftor_{2}$ if it requires fewer resources
for all functionality~$\fun$:
\[
\ftor_{1}\dpleq\ftor_{2}\quad\equiv\quad\ftor_{1}(\fun)\posleq_{\Aressp}\ftor_{2}(\fun),\ \text{for all }\fun\in\funsp.
\]
\end{defn}
\begin{figure}[H]
\begin{centering}
\includegraphics[scale=0.33]{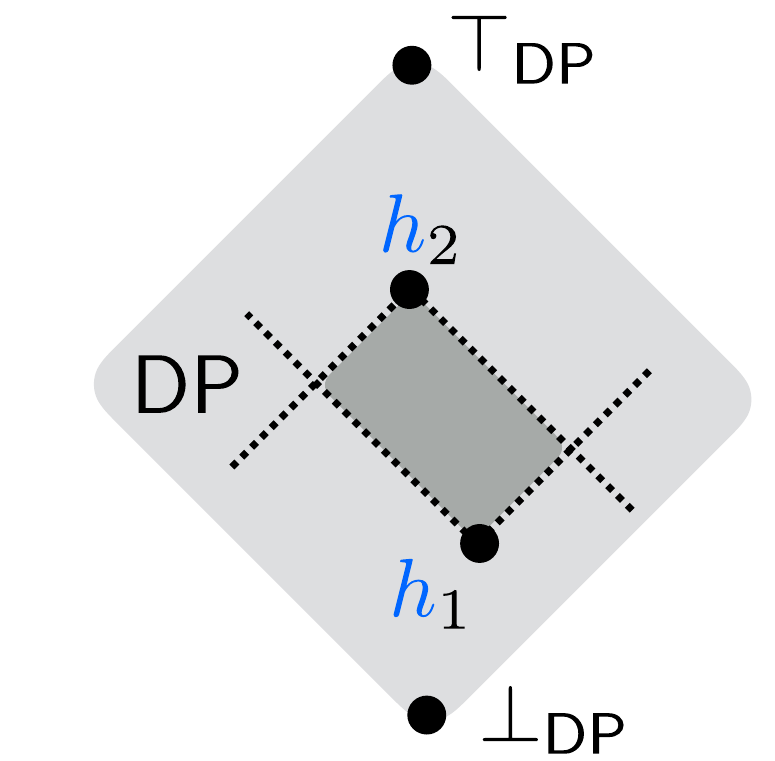}\includegraphics[scale=0.33]{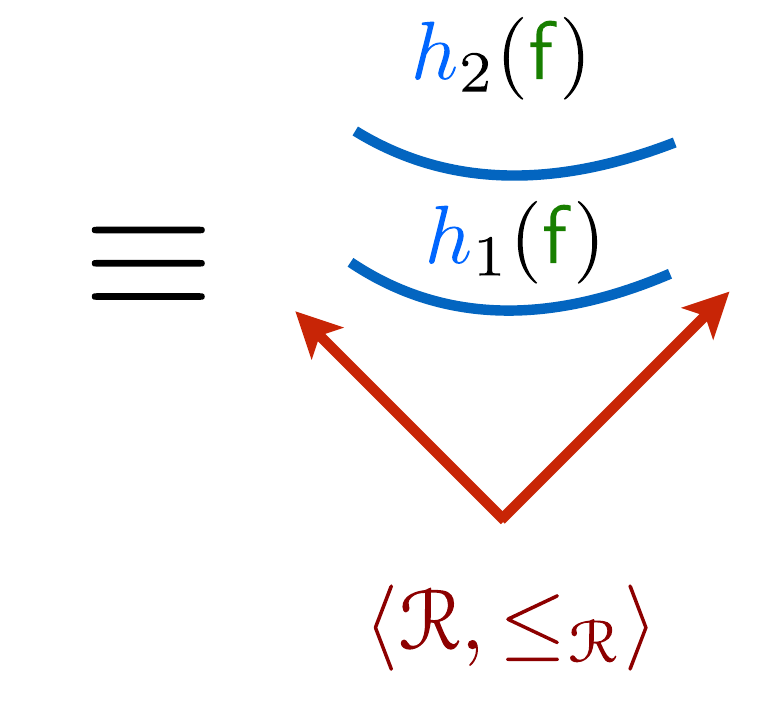}
\par\end{centering}
\caption{The partial order~$\dpleq$ in the space of design problems.}
\end{figure}

In this partial order, there is both a top~$\top_{\dpsp}$ and a
bottom~$\bot_{\dpsp}$, defined as follows:

\vspace{-5mm}

\begin{minipage}[t]{0.4\columnwidth}
\begin{align*}
\bot_{\dpsp}:\funsp & \rightarrow\Aressp,\\
\fun & \mapsto\{\bot_{\ressp}\}.
\end{align*}

\end{minipage}
\begin{minipage}[t]{0.4\columnwidth}
\begin{align}
\top_{\dpsp}:\funsp & \rightarrow\Aressp,\nonumber \\
\fun & \mapsto\emptyset.\label{eq:top}
\end{align}

\end{minipage}

\smallskip{}

$\bot_{\dpsp}$~means that any functionality can be done with zero
resources, and~$\top_{\dpsp}$ means that the problem is always infeasible
(``the set of feasible resources is empty'').

\subsection{Uncertain DPs (UDPs)}
\begin{defn}[Uncertain DPs]
An Uncertain DP (UDP)~$\boldsymbol{u}$ is a pair of DPs~$\langle\udpL\boldsymbol{u},\udpU\boldsymbol{u}\rangle$
such that~$\udpL\boldsymbol{u}\dpleq\udpU\boldsymbol{u}$ .
\end{defn}

\begin{figure}[H]
\begin{centering}
\includegraphics[scale=0.33]{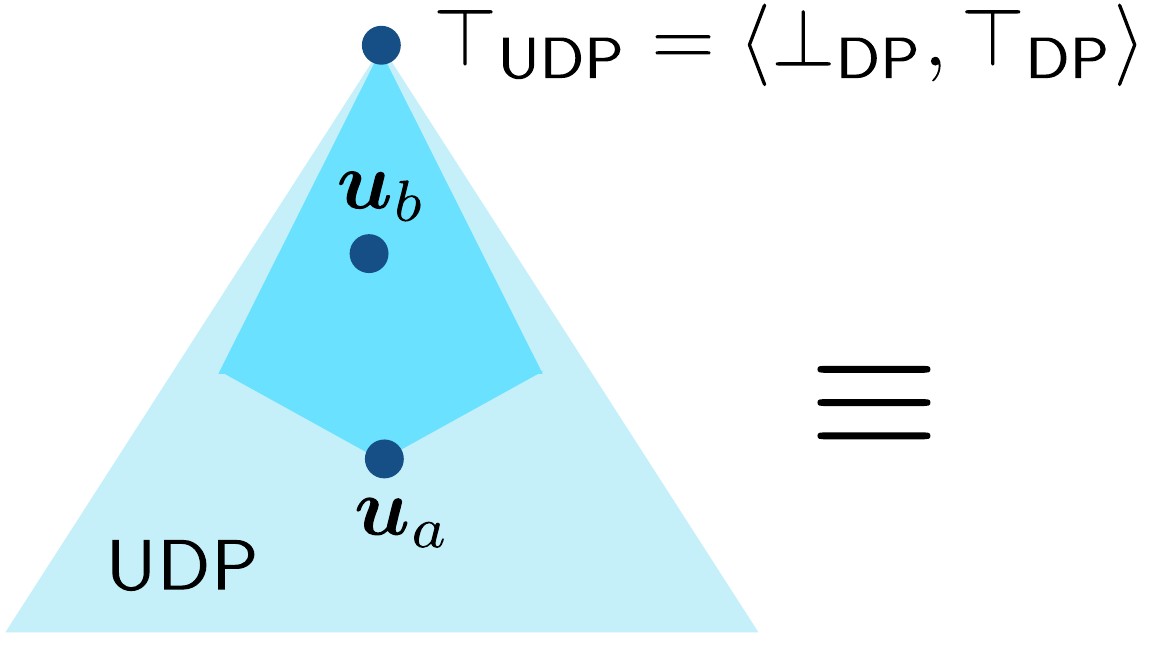}\includegraphics[scale=0.33]{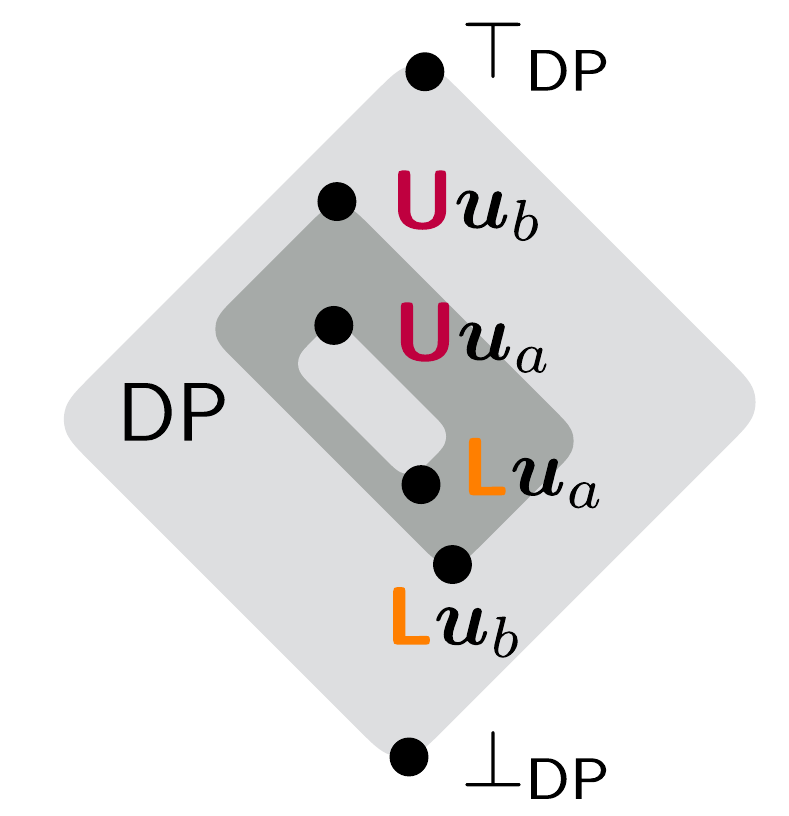}
\par\end{centering}
\caption{\label{fig:udpspace}The partial order~$\udpleq$ has a top~$\top_{\udpsp}=\left\langle \bot_{\dpsp},\top_{\dpsp}\right\rangle .$
This pair describes the state of maximum uncertainty about the DP:
we do not know if the DP is feasible with 0 resources~($\bot_{\dpsp}$),
or if it is completely infeasible~($\top_{\dpsp}$).}
\end{figure}

\begin{defn}[Partial order $\udpleq$]
A UDP~$\udpa$ precedes another UDP~$\udpb$ if the interval~$[\udpL\udpa,\udpU\udpa]$
is contained in the interval~$[\udpL\udpb,\udpU\udpb]$ (\prettyref{fig:udpspace}):
\[
\udpa\udpleq\udpb\quad\equiv\quad\udpL\udpb\dpleq\udpL\udpa\dpleq\udpU\udpa\dpleq\udpU\udpb.
\]
\end{defn}

A DP~$\ftor$ is equivalent to a degenerate UDP~$\langle\ftor,\ftor\rangle$.

A UDP~$\boldsymbol{u}$ is a bound for a DP~$\ftor$ if~$\boldsymbol{u}\udpleq\langle\ftor,\ftor\rangle$,
or, equivalently, if $\udpL\boldsymbol{u}\udpleq\ftor\udpleq\udpU\boldsymbol{u}$.

\section{Interconnection of Uncertain Design Problems\label{sec:UMCDP}}

We now define the interconnection of UDPs, in an equivalent way to
the definition of MCDPs. The only difference between \prettyref{def:MCDP-algebraic}
and~\prettyref{def:umcdp} below is that the valuation assigns to
each atom an UDP, rather than a~DP.
\begin{defn}[Algebraic definition of UMCDPs]
\label{def:umcdp}An Uncertain MCDP (UMCDP) is a tuple~$\left\langle \atoms,\atree,\val\right\rangle $,
where~$\atoms$ is a set of atoms,~$\atree\in\terms(\{\dpseries,\dppar,\dploop\},\atoms)$
is the algebraic representation of the graph, and~$\val:\atoms\rightarrow\udpsp$
is a valuation that assigns to each atom a UDP.
\end{defn}

Next, the semantics of a UMCDP is defined as a map~$\udpsem$ that
computes the UDP. \prettyref{def:semantics-udp}~below is analogous
to~\prettyref{def:dpsem}.
\begin{defn}[Semantics of UMCDPs]
\label{def:semantics-udp}Given an UMCDP~$\left\langle \atoms,\atree,\val\right\rangle $,
the semantics function~$\udpsem$ computes a UDP
\[
\udpsem\llbracket\left\langle \atoms,\atree,\val\right\rangle \rrbracket\in\udpsp,
\]
and it is recursively defined as follows:

\adjustbox{max width=8.6cm}{
\noindent\begin{minipage}[t]{1\columnwidth}
\[
\udpsem\llbracket\left\langle \atoms,a,\val\right\rangle \rrbracket=\val(a),\qquad\text{for all}\ a\in\atoms.
\]
\begin{align*}
\udpL\udpsem\llbracket\left\langle \atoms,\dpseries(\atree_{1},\atree_{2}),\val\right\rangle \rrbracket & =(\udpL\udpsem\llbracket\left\langle \atoms,\atree_{1},\val\right\rangle \rrbracket)\,\opseries\,(\udpL\udpsem\llbracket\left\langle \atoms,\atree_{2},\val\right\rangle \rrbracket),\\
\udpU\udpsem\llbracket\left\langle \atoms,\dpseries(\atree_{1},\atree_{2}),\val\right\rangle \rrbracket & =(\udpU\udpsem\llbracket\left\langle \atoms,\atree_{1},\val\right\rangle \rrbracket)\,\opseries\,(\udpU\udpsem\llbracket\left\langle \atoms,\atree_{2},\val\right\rangle \rrbracket),
\end{align*}
\begin{align*}
\udpL\udpsem\llbracket\left\langle \atoms,\dppar(\atree_{1},\atree_{2}),\val\right\rangle ] & =(\udpL\udpsem\llbracket\left\langle \atoms,\atree_{1},\val\right\rangle \rrbracket)\ \oppar\ (\udpL\udpsem\llbracket\left\langle \atoms,\atree_{2},\val\right\rangle \rrbracket),\\
\udpU\udpsem\llbracket\left\langle \atoms,\dppar(\atree_{1},\atree_{2}),\val\right\rangle ] & =(\udpU\udpsem\llbracket\left\langle \atoms,\atree_{1},\val\right\rangle \rrbracket)\ \oppar\ (\udpU\udpsem\llbracket\left\langle \atoms,\atree_{2},\val\right\rangle \rrbracket),
\end{align*}
\begin{align*}
\udpL\udpsem\llbracket\left\langle \atoms,\dploop(\atree),\val\right\rangle \rrbracket & =(\udpL\udpsem\llbracket\left\langle \atoms,\atree,\val\right\rangle \rrbracket)^{\oploop},\\
\udpU\udpsem\llbracket\left\langle \atoms,\dploop(\atree),\val\right\rangle \rrbracket & =(\udpU\udpsem\llbracket\left\langle \atoms,\atree,\val\right\rangle \rrbracket)^{\oploop}.
\end{align*}

\end{minipage}}
\end{defn}
The operators $\oploop,\opseries,\oppar$ are defined in \prettyref{def:opseries}\textendash \prettyref{def:oploop}.

\section{Approximation results\label{sec:Approximation-results}}

The main result of this section is a relaxation result stated as \prettyref{thm:udpsem-monotone}
below. The following is an informal statement. 

\begin{figure}[b]
\begin{centering}
\subfloat[\label{fig:consider1}Consider an MCDP containing the DP $\ftor_{a}$.]{\begin{centering}
\minwidthbox{\includegraphics[scale=0.33]{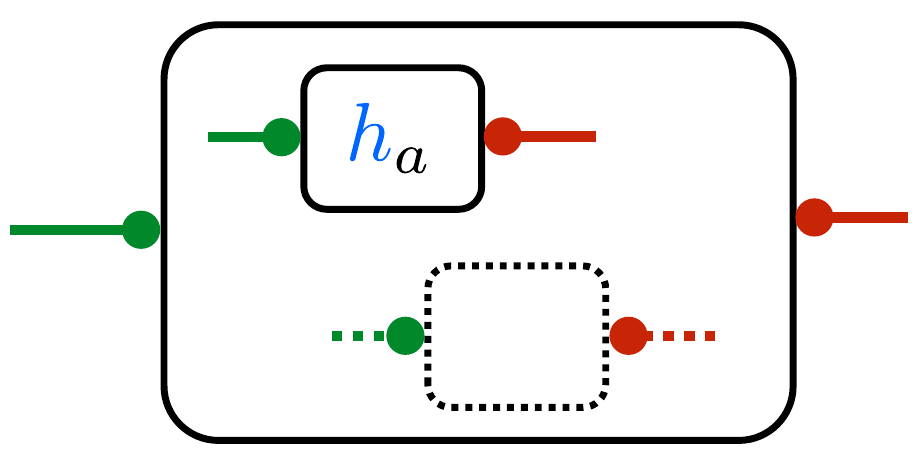}}{8cm}
\par\end{centering}
}\\
\subfloat[\label{fig:consider2}Suppose that a bound $\left\langle \boldsymbol{\mathsf{L}},\boldsymbol{\mathsf{U}}\right\rangle $
is known.]{\begin{centering}
\minwidthbox{\includegraphics[scale=0.33]{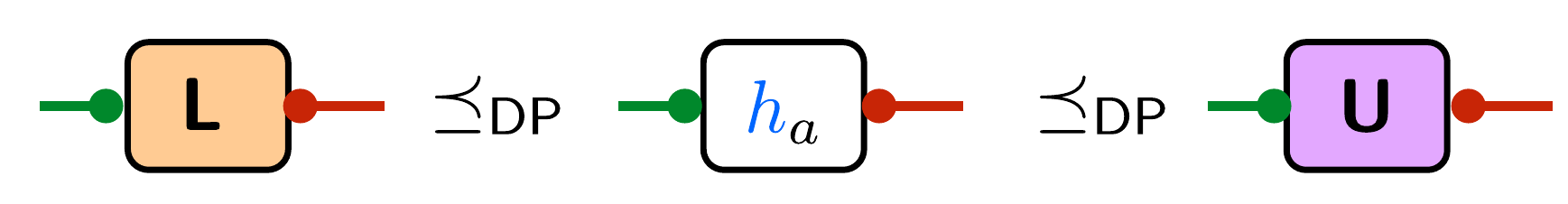}}{8cm}
\par\end{centering}
}\\
\subfloat[\label{fig:luinside}Replace the DP $\ftor_{a}$ by the uncertain
bounds. ]{\begin{centering}
\minwidthbox{\includegraphics[scale=0.33]{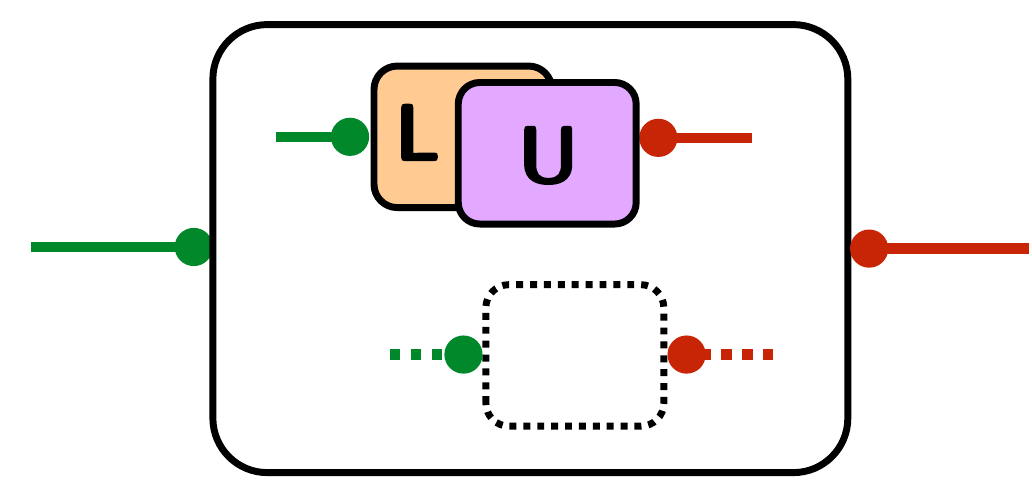}}{8cm}
\par\end{centering}
}\\
\subfloat[\label{fig:pair}Create a pair of MCDPs by choosing either lower or
upper bound. ]{\begin{centering}
\minwidthbox{\includegraphics[scale=0.25]{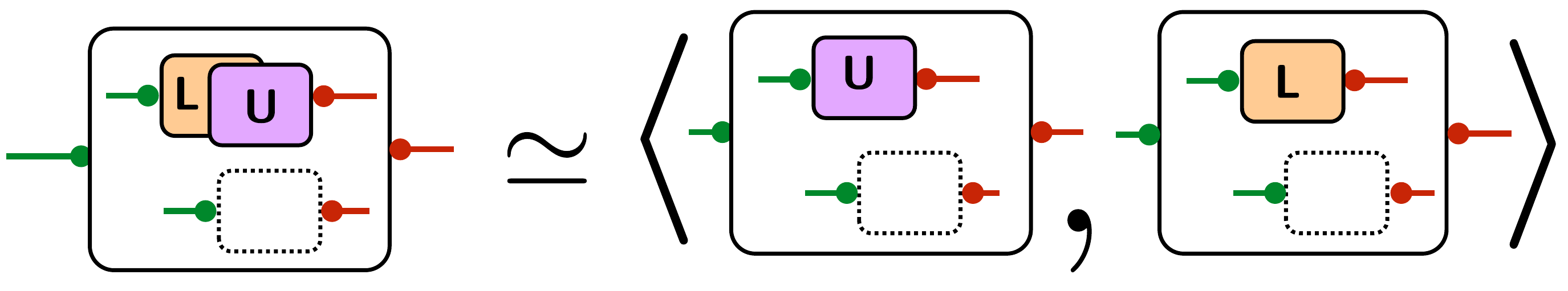}}{8cm}
\par\end{centering}
}\\
\subfloat[\label{fig:domin}The resulting MCDPs are bounds for the original
MCDP. ]{\begin{centering}
\minwidthbox{\includegraphics[scale=0.33]{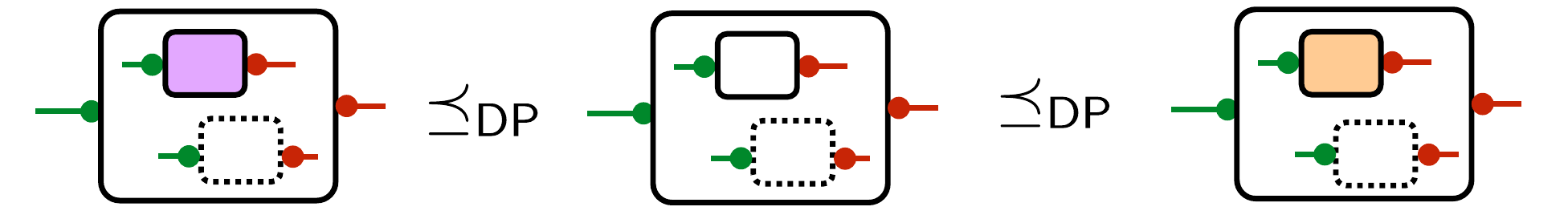}}{8cm}
\par\end{centering}
}
\par\end{centering}
\caption{Illustrations for the informal statement of the main result.}
\end{figure}

\subsubsection*{Informal statement}

Consider an MCDP composed of many DPs, and call one~$\ftor_{a}$~(\prettyref{fig:consider1}).
Suppose there exist two DPs~$\boldsymbol{\mathsf{L}}$, $\boldsymbol{\mathsf{U}}$
that bound the DP~$\ftor_{a}$ in the sense that~$\boldsymbol{\mathsf{L}}\dpleq\ftor_{a}\dpleq\boldsymbol{\mathsf{U}}$~(\prettyref{fig:consider2}).
This can model either uncertainty in our knowledge of~$\ftor_{a}$,
or it can be a relaxation that we willingly introduce. The pair~$\left\langle \boldsymbol{\mathsf{L}},\boldsymbol{\mathsf{U}}\right\rangle $
forms a UDP that can be plugged in the MCDP in place of~$\ftor_{a}$
(\prettyref{fig:luinside}). We will see that if we plug in only~$\boldsymbol{\mathsf{L}}$
or~$\boldsymbol{\mathsf{U}}$ separately in place of the original
DP~$\ftor_{a}$~(\prettyref{fig:pair}), we obtain a pair of MCDPs
that form a UDP. Moreover, the solution of these two MCDPs are upper
and lower bounds for the solution of the original MCDP~(\prettyref{fig:domin}).
Therefore, we can propagate the uncertainty in the model to the uncertain
in the solution. This result generalizes for any number of substitutions.

\subsubsection*{Formal statement}

First, we define a partial order on the valuations. A valuation precedes
another if it gives more information on each DP.
\begin{defn}[Partial order $\posleq_{V}$ on valuations]
\label{def:For-two-valuations,}For two valuations~$\val_{1},\val_{2}:\atoms\rightarrow\udpsp$,
say that~$\val_{1}\posleq_{V}\val_{2}$ if~$\val_{1}(a)\udpleq\val_{2}(a)$
for all~$a\in\atoms$.
\end{defn}
At this point, we have enough machinery in place that we can simply
state the result as ``the semantics is monotone in the valuation''.
\begin{thm}[$\udpsem$ is monotone in the valuation]
\label{thm:udpsem-monotone}If $\val_{1}\posleq_{V}\val_{2}$, then
\[
\udpsem\llbracket\left\langle \atoms,\atree,\val_{1}\right\rangle \rrbracket\udpleq\udpsem\llbracket\left\langle \atoms,\atree,\val_{2}\right\rangle \rrbracket.
\]
\end{thm}
\begin{IEEEproof}
\C{This follows easily from the definitions in \prettyref{def:For-two-valuations,}.
As intermediate results, first prove that the lower bound~$\udpL\udpsem$
is monotone in the valuation with respect to the order~$\dpleq$:
\[
\udpL\udpsem\llbracket\langle\atoms,\atree,\val_{1}\rangle\rrbracket\dpleq\udpL\udpsem\llbracket\langle\atoms,\atree,\val_{2}\rangle\rrbracket.
\]
Then repeat the same reasoning for~$\udpU$, to obtain: 
\[
\udpU\udpsem\llbracket\langle\atoms,\atree,\val_{2}\rangle\rrbracket\dpleq\udpU\udpsem\llbracket\langle\atoms,\atree,\val_{1}\rangle\rrbracket.
\]
These two together allow to conclude that $\udpsem$ is monotone with
respect to the valuation with respect to the order~$\udpleq$:}
\[
\udpsem\llbracket\langle\atoms,\atree,\val_{1}\rangle\rrbracket\udpleq\udpsem\llbracket\langle\atoms,\atree,\val_{2}\rangle\rrbracket.
\]
\end{IEEEproof}

This result says that we can swap any DP in a MCDP with a UDP relaxation
to obtain a UMCDP, which then we can solve to obtain inner and outer
approximations to the solution of the original MCDP. This shows that
considering interval uncertainty in the MCDP framework is easy because
it reduces to solving a pair of problems instead of one. The rest
of the paper consists of applications of this result.

\section{Applications\label{sec:Applications}}

This section shows three example applications of the theory:
\begin{enumerate}
\item The first example deals with \emph{parametric uncertainty}.
\item The second example deals with the idea of relaxation of a scalar relation.
This is equivalent to accepting a \emph{tolerance} for a given variable,
in exchange for reduced computation.
\item The third example deals with the\emph{ relaxation of relations with
infinite cardinality}. In particular it shows how one can obtain consistent
estimates with a finite and prescribed amount of computation.
\end{enumerate}

\subsection{Parametric Uncertainty\label{sec:Application-uncertainty}}

To instantiate the model in~\prettyref{fig:Example1}, we need to
obtain numbers for energy density, specific cost, and operating life
for all batteries technologies we want to examine. By browsing Wikipedia,
one can find the figures in~\prettyref{tab:batteries}. 

\begin{table}[H]
\begin{centering}
\caption{\label{tab:batteries}Specifications of common batteries technologies}
\par\end{centering}
\centering{}{\footnotesize{}}
\begin{tabular}{crr@{\extracolsep{0pt}.}lr}
\multirow{2}{*}{{\footnotesize{}\tableColors}\emph{\footnotesize{}technology}} & \emph{\footnotesize{}energy density} & \multicolumn{2}{c}{\emph{\footnotesize{}specific cost}} & \emph{\footnotesize{}operating life}\tabularnewline
 & {\footnotesize{}{[}Wh/kg{]}} & \multicolumn{2}{c}{{\footnotesize{}{[}Wh/\${]}}} & \# cycles\tabularnewline
{\footnotesize{}NiMH} & {\footnotesize{}100} & {\footnotesize{}3}&{\footnotesize{}41 } & {\footnotesize{}500 }\tabularnewline
{\footnotesize{}NiH2} & {\footnotesize{}45} & {\footnotesize{}10}&{\footnotesize{}50 } & {\footnotesize{}20000}\tabularnewline
{\footnotesize{}LCO} & {\footnotesize{}195} & {\footnotesize{}2}&{\footnotesize{}84} & {\footnotesize{}750}\tabularnewline
{\footnotesize{}LMO} & {\footnotesize{}150} & {\footnotesize{}2}&{\footnotesize{}84 } & {\footnotesize{}500}\tabularnewline
{\footnotesize{}NiCad} & {\footnotesize{}30} & {\footnotesize{}7}&{\footnotesize{}50 } & {\footnotesize{}500}\tabularnewline
{\footnotesize{}SLA} & {\footnotesize{}30} & {\footnotesize{}7}&{\footnotesize{}00} & {\footnotesize{}500}\tabularnewline
{\footnotesize{}LiPo} & {\footnotesize{}150} & {\footnotesize{}2}&{\footnotesize{}50} & {\footnotesize{}600}\tabularnewline
{\footnotesize{}LFP} & {\footnotesize{}90} & {\footnotesize{}1}&{\footnotesize{}50} & {\footnotesize{}1500}\tabularnewline
\end{tabular}{\footnotesize \par}
\end{table}

Should we trust those figures? Fortunately, we can easily deal with
possible mistrust by introducing uncertain DPs. Formally, we replace
the DPs for\emph{ energy density}, \emph{specific cost}, \emph{operating
life} in~\prettyref{fig:Example1} with the corresponding Uncertain
DPs with a configurable uncertainty. We can then solve the UDPs to
obtain a lower bound and an upper bound to the solutions that can
be presented to the user.

\prettyref{fig:unc_battery_uncertain} shows the relation between
the provided \F{endurance} and the minimal \R{total mass} required,
when using uncertainty of $5\%$, $10\%$, $25\%$ on the numbers
above. Each panel shows two curves: the lower bound (best case analysis)
and the upper bound (worst case analysis). In some cases, the lower
bound is feasible, but the upper bound is not. For example, in panel~\emph{b},
for 10\% uncertainty, we can conclude that, notwithstanding the uncertainty,
there exists a solution for endurance~$\leq1.35\,\text{hours}$,
while for values in $[1.35,1.5]$, represented by the shaded area,
we cannot conclude either the existence or nonexistence of a solution,
because the lower bound is feasible, but the upper bound is not. This
area of uncertainty depends on the parameter uncertainty injected
(smaller for 5\%, and larger for 25\%). 
\begin{center}
\begin{figure}[H]
\begin{centering}
\includegraphics[scale=0.33]{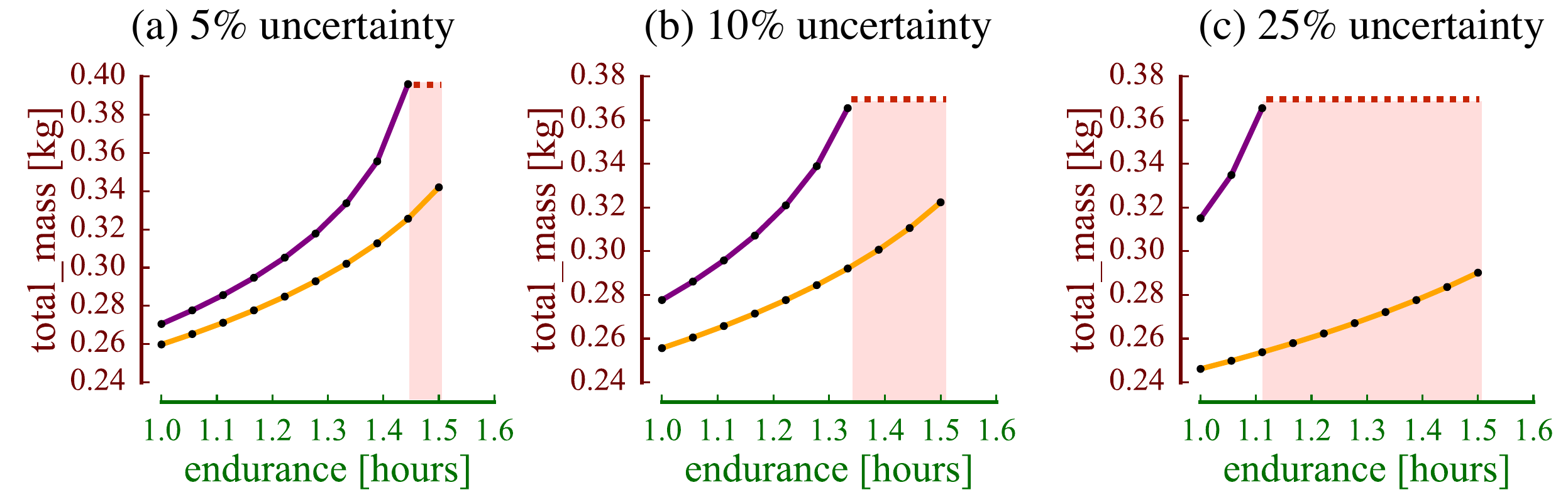}
\par\end{centering}
\caption{\label{fig:unc_battery_uncertain}Uncertain relation between \F{endurance}
and the minimal \R{total mass} required, obtained by solving the
example in \prettyref{fig:Example1} for different values of the uncertainty
on the characteristics of the batteries (5\%, 10\%, 25\%). The shaded
area represents the subset of the functionality space for which we
cannot conclude that a solution exists, because the upper bound DP
is not feasible, while the lower bound DP is feasible. }
\end{figure}
\par\end{center}

\subsection{Introducing Tolerances\label{sec:Application-tolerance}}

Another application of the theory is the introduction of tolerances
for any variable in the optimization problem. For example, one might
not care about the variations of the battery mass below, say,~$1\,\text{g}$.
One can then introduce a $\pm1\,\text{\ensuremath{\text{g}} }$ uncertainty
in the definition of the problem by adding a UDP hereby called ``uncertain
identity''.

\subsubsection{The uncertain identity}

\begin{figure}[H]
\subfloat{\centering{}\includegraphics[scale=0.33]{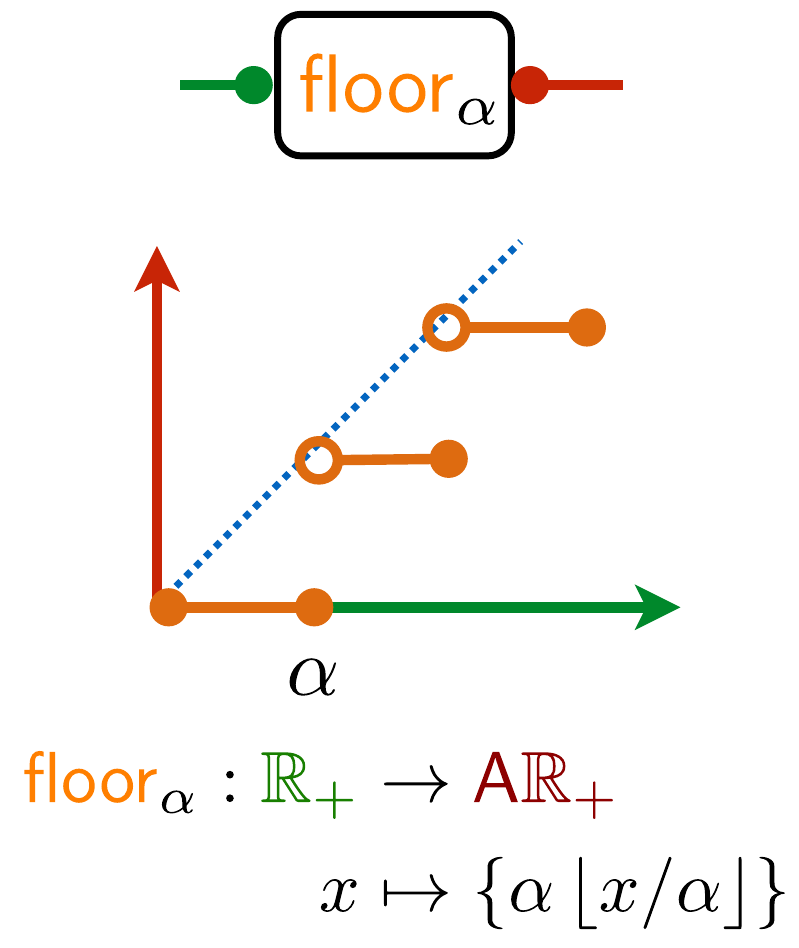}}\hfill{}\subfloat{\includegraphics[scale=0.33]{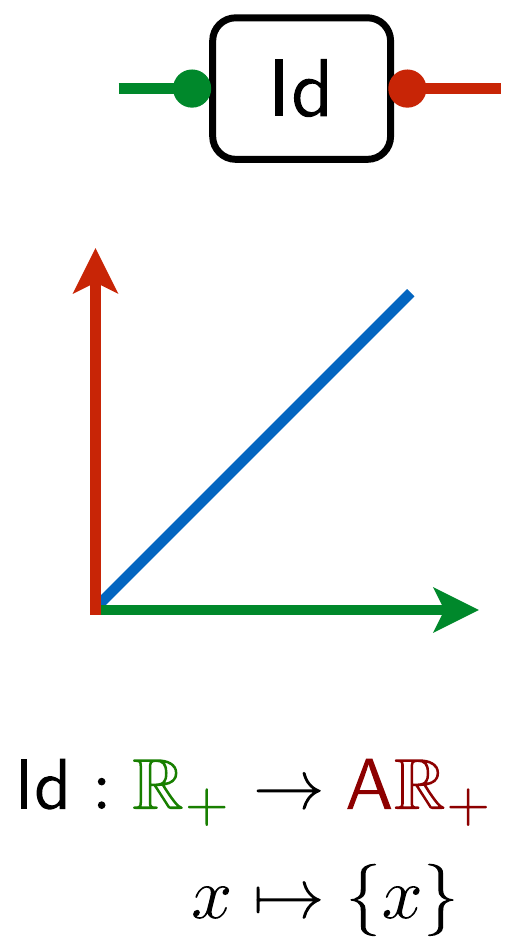}}\hfill{}\subfloat{\includegraphics[scale=0.33]{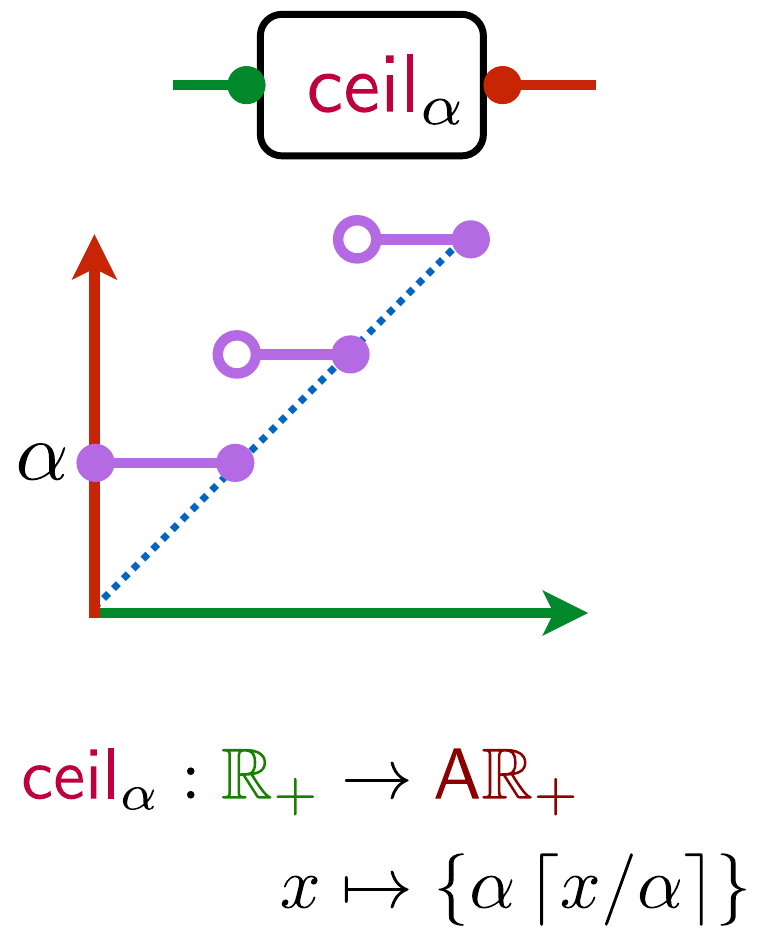}}

\caption{\label{fig:identity_approximation}The identity and its two relaxations~$\ufloor_{\alpha}$
and $\uceil_{\alpha}$.}
\end{figure}

Let~$\alpha>0$ be a step size. Define~$\ufloor_{\alpha}$ and~$\uceil_{\alpha}$
to be the floor and ceil with step size~$\alpha$~(\prettyref{fig:identity_approximation}).
By construction, $\ufloor_{\alpha}\dpleq\mathsf{Id}\dpleq\uceil_{\alpha}.$ 

\begin{figure}[H]
\begin{centering}
\includegraphics[scale=0.33]{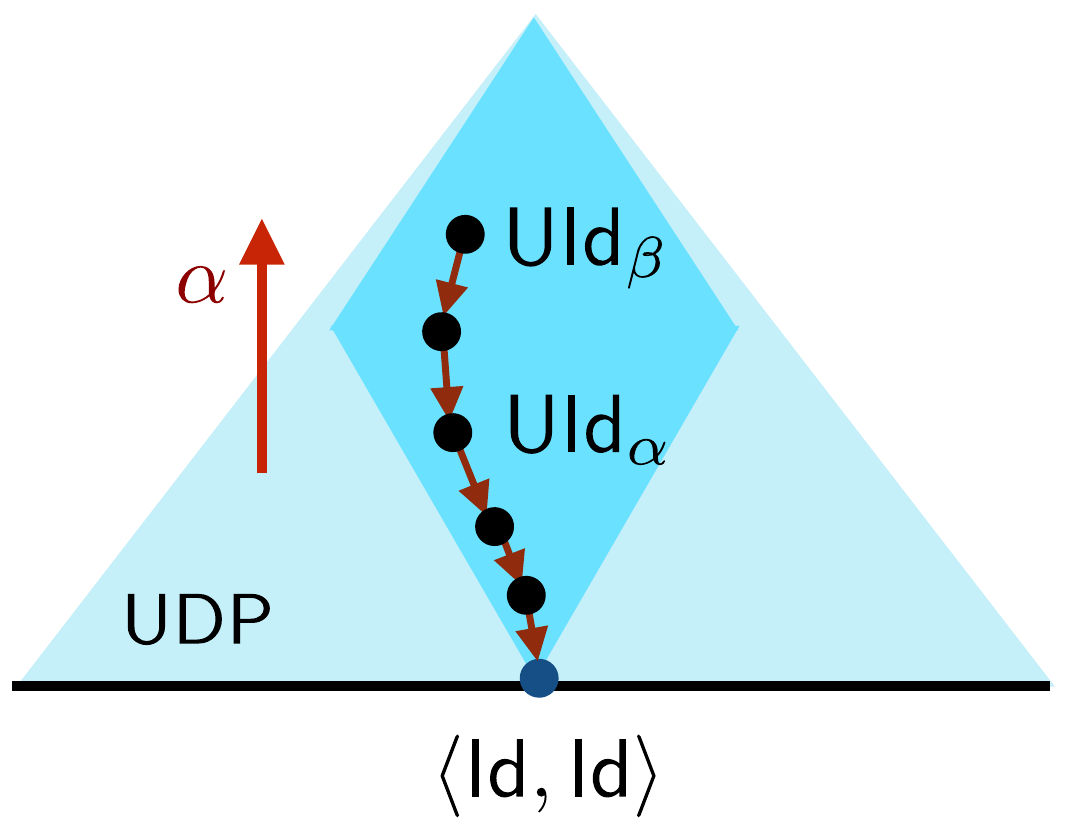}\includegraphics[scale=0.33]{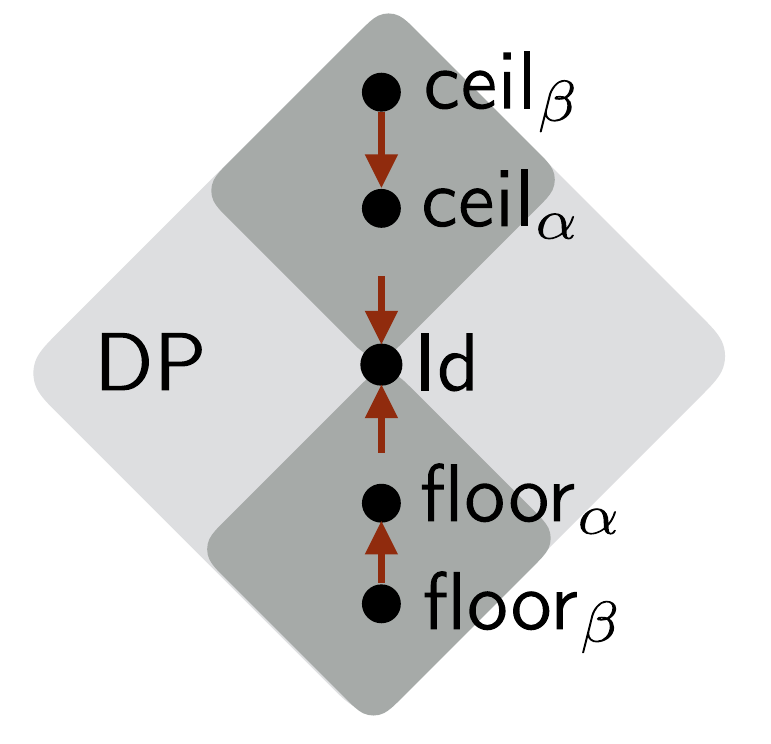}
\par\end{centering}
\caption{\label{fig:other}Convergence of $\UId_{\alpha}$ to the identity~$\mathsf{Id}$. }
\end{figure}

Let $\UId_{\alpha}\doteq\left\langle \ufloor_{\alpha},\uceil_{\alpha}\right\rangle $
be the ``uncertain identity''. For~$0<\alpha<\beta$, it holds
that
\[
\mathsf{Id}\poslt_{\udpsp}\UId_{\alpha}\poslt_{\udpsp}\UId_{\beta}.
\]
Therefore, the sequence $\UId_{\alpha}$ is a descending chain that
converges to~$\mathsf{Id}$ as~$\alpha\rightarrow0$~(\prettyref{fig:other}).

\subsubsection{Approximations in MCDP}

\begin{figure}[H]
\centering{}\includegraphics[scale=0.33]{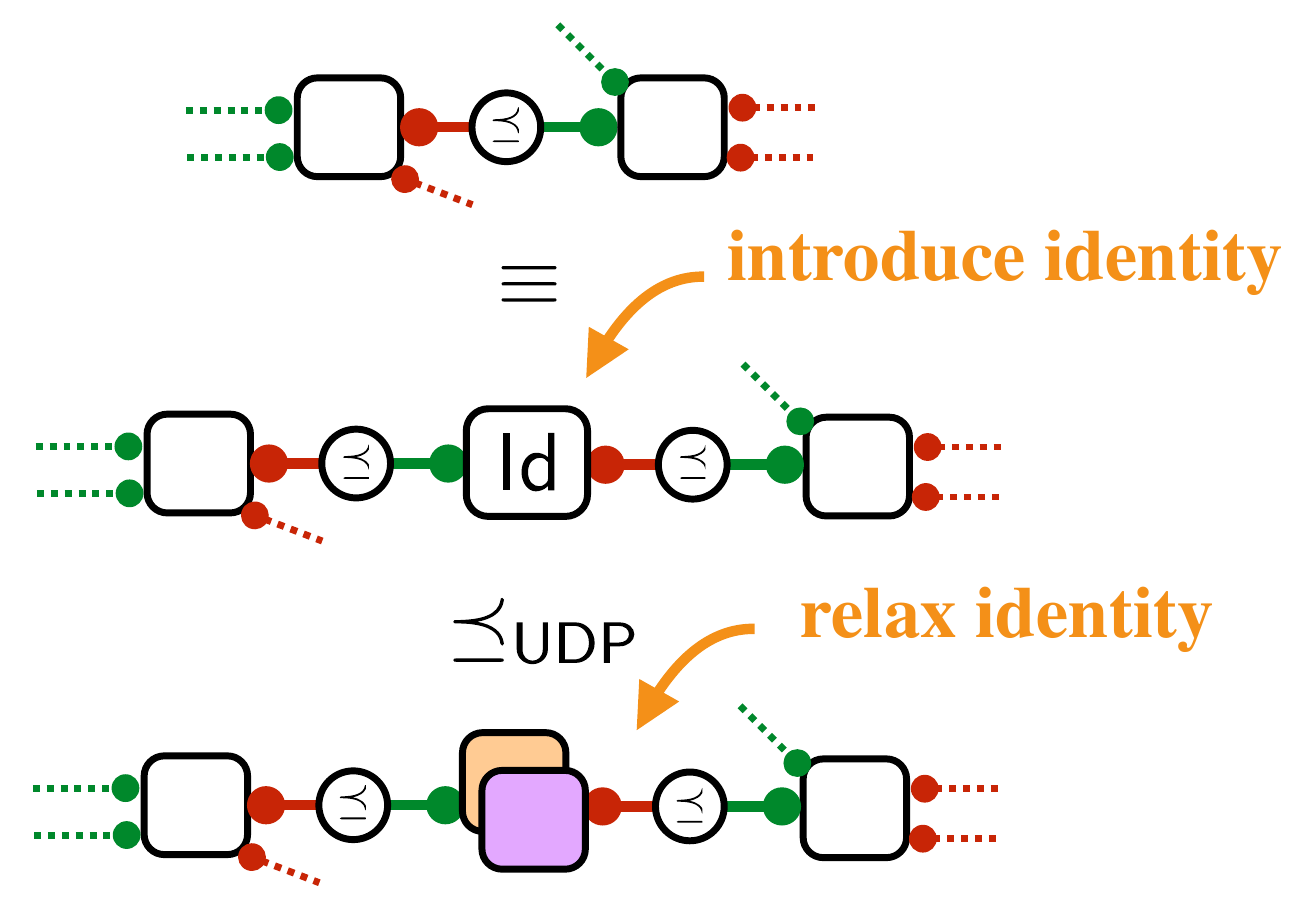}\caption{\label{fig:introduce}Introduction of tolerances by using the ``uncertain
identity'' $\text{UId}_{\alpha}$.}
\end{figure}

We can take any edge in an MCDP and apply this relaxation. Formally,
we first introduce an identity~$\mathsf{Id}$ and then relax it using~$\UId_{\alpha}$~(\prettyref{fig:introduce}).

Mathematically, given an MCDP~$\left\langle \atoms,\atree,\val\right\rangle $,
we generate a UMCDP~$\left\langle \atoms,\atree,\val_{\alpha}\right\rangle $,
where the new valuation~$\val_{\alpha}$ agrees with~$\val$ except
on a particular atom~$a\in\atoms$, which is replaced by the series
of the original~$\val(a)$ and the approximation~$\text{UId}_{\alpha}$:
\begin{align*}
\val_{\alpha}(a) & \doteq\dpseries(\UId_{\alpha},\val(a))
\end{align*}
Call the original and approximated DPs~$\dprob$ and~$\dprob_{\alpha}$:
\[
\begin{array}{ccc}
\dprob\doteq\udpsem\left\llbracket \left\langle \atoms,\atree,\val\right\rangle \right\rrbracket , &  & \dprob{}_{\alpha}\doteq\udpsem\left\llbracket \left\langle \atoms,\atree,\val_{\alpha}\right\rangle \right\rrbracket .\end{array}
\]
Because $\val\posleq_{V}\val_{\alpha}$ (in the sense of~\prettyref{def:For-two-valuations,}),
\prettyref{thm:udpsem-monotone} implies that 
\[
\dprob\udpleq\dprob_{\alpha}.
\]
This means that we can solve~$\udpL\dprob_{\alpha}$ and~$\udpU\dprob_{\alpha}$
and obtain upper and lower bounds for~$\dprob$. Furthermore, by
varying~$\alpha$, we can construct an approximating sequence of
DPs whose solution will converge to the solution of the original MCDP.

\paragraph*{Numerical results}

\begin{figure}[H]
\subfloat[\label{fig:mass}]{\includegraphics[scale=0.4]{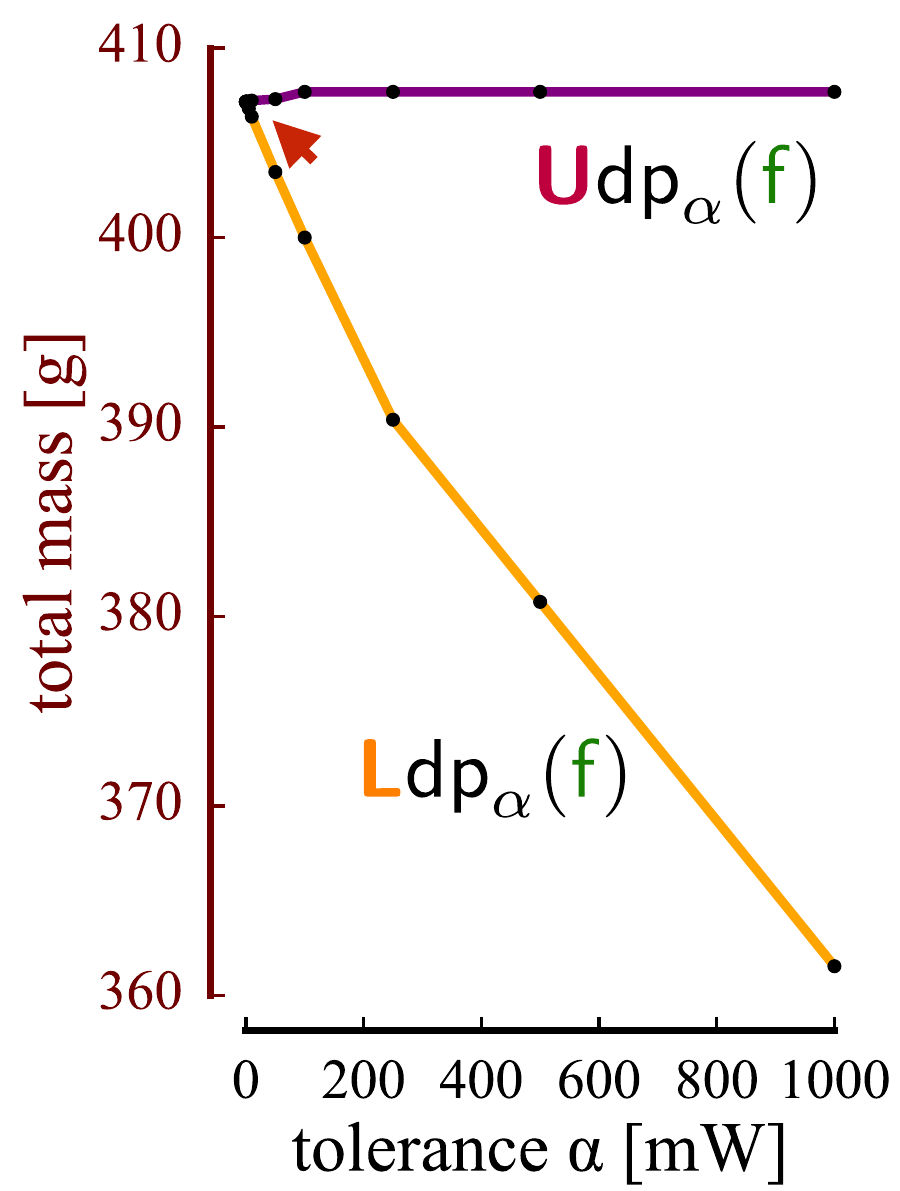}

}\subfloat[\label{fig:num_iterations}]{\includegraphics[scale=0.4]{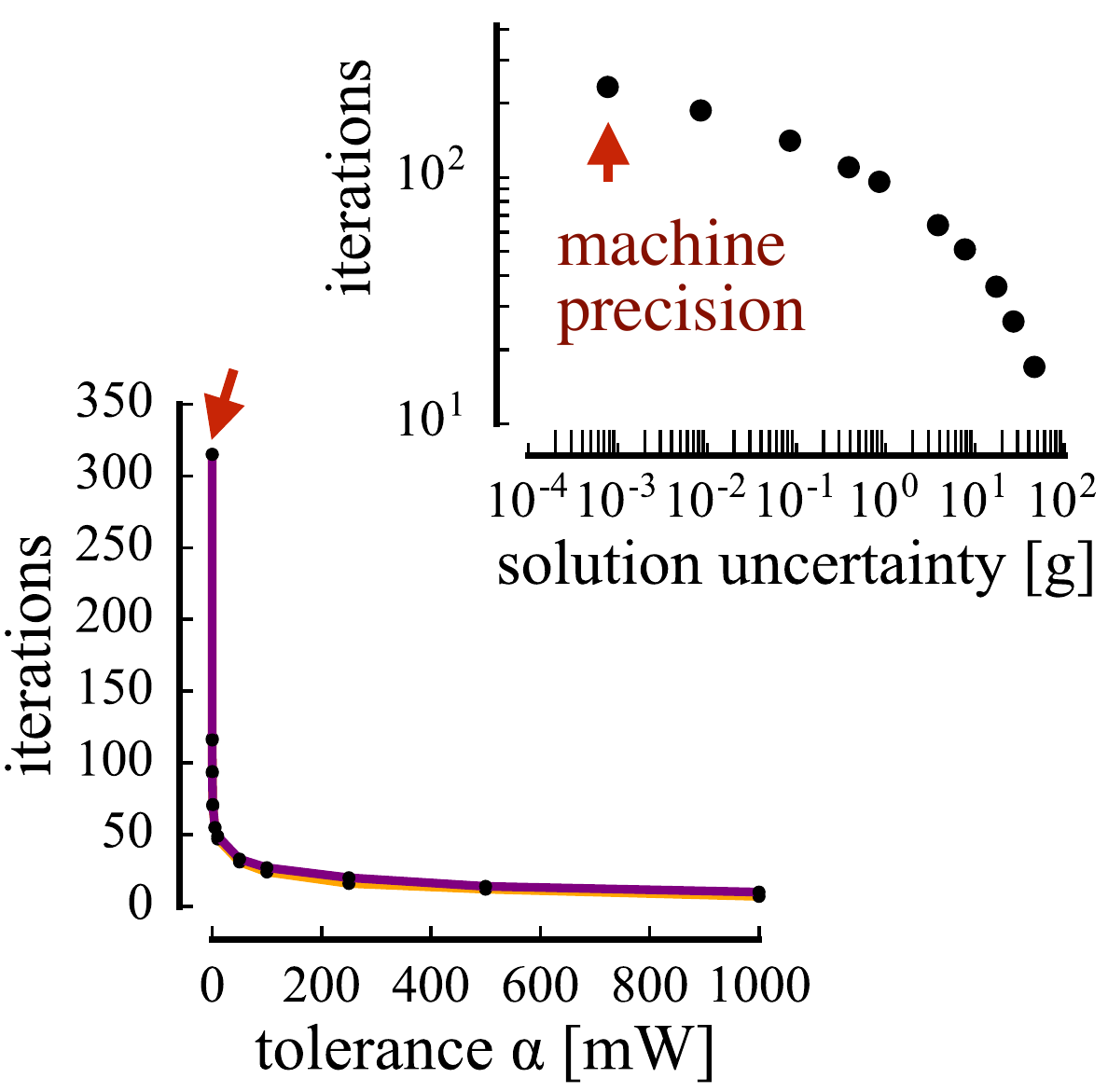}

}

\caption{Results of model in \prettyref{fig:Example1} when tolerance is applied
to the actuation \R{power} resource. Panel~\emph{a} shows the uncertainty
in the solution as a function of tolerance. Panel~\emph{b} shows
the iterations needed.}
\end{figure}

This procedure was applied to the example model in~\prettyref{fig:Example1}
by introducing a tolerance to the ``power'' variable for the actuation.
The tolerance~$\alpha$ is chosen at logarithmic intervals between~$0.01\,\text{mW}$
and~$1\,\text{W}$. \prettyref{fig:mass}~shows the solutions of
the minimal mass required for~$\udpL\dprob_{\alpha}$ and~$\udpU\dprob_{\alpha}$,
as a function of~$\alpha$. \prettyref{fig:mass} confirms the consistency
results predicted by the theory. First, if the solutions for both~$\udpL\dprob_{\alpha}$
and~$\udpU\dprob_{\alpha}$ exist, then they are ordered ($\udpL\dprob_{\alpha}(\fun)\posleq\udpU\dprob_{\alpha}(\fun)$).
Second, as~$\alpha$ decreases, the interval shrinks. Third, the
bounds are consistent, in the sense that the solution for the original
DP is always contained in the bounds.

Next, it is interesting to consider the computational complexity.
\prettyref{fig:num_iterations}~shows the number of iterations as
a function of the resolution~$\alpha$, and the trade-off of the
uncertainty of the solution and the computational resources spent.
This shows that this approximation scheme is an effective way to reduce
the computation load while maintaining a consistent estimate.

\subsection{Relaxation for relations with antichains of infinite cardinality\label{sec:Application-relax}}

Another way in which uncertain DPs can be used is to construct approximations
of DPs that would be too expensive to solve exactly. For example,
consider a relation like 
\begin{equation}
{\colF\text{travel\_distance}}\leq{\colR\text{velocity}}\times{\colR\text{endurance}},\label{eq:qun}
\end{equation}
which appears in the model in \prettyref{fig:Example1}. If we take
these three quantities in \prettyref{eq:qun} as belonging to~$\reals$,
then, for each value of the \F{travel distance}, there are infinite
pairs of~$\left\langle {\colR\text{velocity}},{\colR\text{endurance}}\right\rangle $
that are feasible. On a computer, if the quantities are represented
as floating point numbers, the combinations are properly not ``infinite'',
but, still, extremely large. We can avoid considering all combinations
by creating a sequence of uncertain DPs that use finite and prescribed
computation.

\subsubsection{Relaxations for addition}

\begin{figure}[H]
\begin{centering}
\includegraphics[scale=0.33]{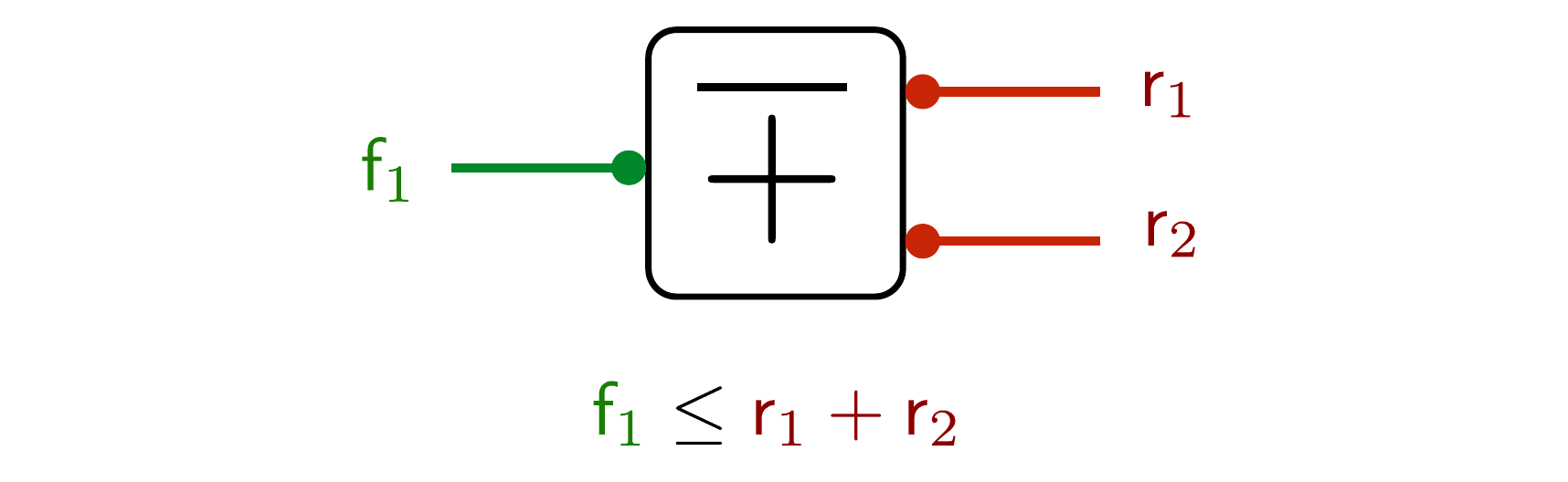}
\par\end{centering}
\caption{\label{fig:example-invplus}Graphical relation of the DP for the relation
$\fun_{1}\leq\res_{1}+\res_{2}$. The overline in ``$\overline{+}$''
represents the fact that this is the ``dual'' of the relation~$\fun_{1}+\fun_{2}\leq\res$,
for which we used the name ``$+$''.}
\end{figure}

Consider a monotone relation between some functionality~$\fun_{1}\in\mathbb{R}_{+}$
and resources~$\res_{1},\res_{2}\in\mathbb{R}_{+}$ described by
the constraint that~$\fun_{1}\leq\res_{1}+\res_{2}$ (\prettyref{fig:example-invplus}).
For example, this could represent the case where there are two batteries
providing the power~$\fun_{1}$, and we need to decide how much to
allocate to the first~($\res_{1}$) or the second~($\res_{2}$).

The formal definition of this constraint as an DP is
\begin{align*}
\overline{+}:{\colF\reals_{+}} & \rightarrow{\colR\antichains(\reals_{+}\times\reals_{+})},\\
\fun_{1} & \mapsto\{\left\langle x,\fun_{1}-x\right\rangle \mid x\in\reals_{+}\}.
\end{align*}
Note that, for each value~$\fun_{1}$, $\overline{+}(\fun_{1})$
is a set of infinite cardinality. We will now define two sequences
of relaxations for~$\overline{+}$ with a fixed number of solutions~$n\geq1$. 

\subsubsection*{Using uniform sampling}

\begin{figure}[H]
\centering{}\includegraphics[bb=0bp 10bp 537bp 157bp,clip,scale=0.33]{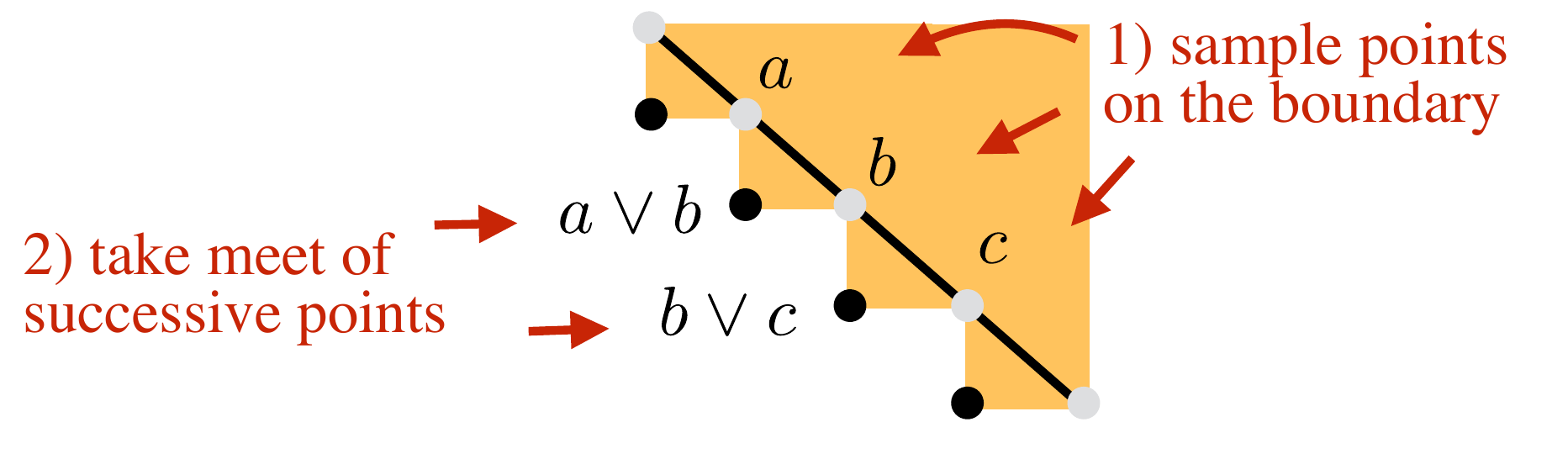}\caption{\label{fig:make_lower}Construction of approximating antichains.}
\end{figure}

We will first define a sequence of UDPs~$S_{n}$ based on uniform
sampling. Let~$\udpU S_{n}$ consist of~$n$ points sampled on the
segment with extrema~$\left\langle 0,\fun_{1}\right\rangle $ and~$\left\langle \fun_{1},0\right\rangle $.
For~$\udpL S_{n}$, sample~$n+1$ points on the segment and take
the \emph{meet} of successive points~(\prettyref{fig:make_lower}). 

The first elements of the sequences are shown in~\prettyref{fig:approx_invplus}.
One can easily prove that $\udpL S_{n}\dpleq\overline{+}\dpleq\udpU S_{n}$,
and thus~$S_{n}$ is a relaxation of~$\overline{+}$, in the sense
that~$\overline{+}\udpleq S_{n}$. Moreover, $S_{n}$ converges to
$\overline{+}$ as $n\rightarrow\infty$. 
\begin{center}
\begin{figure}[H]
\centering{}\includegraphics[scale=0.33]{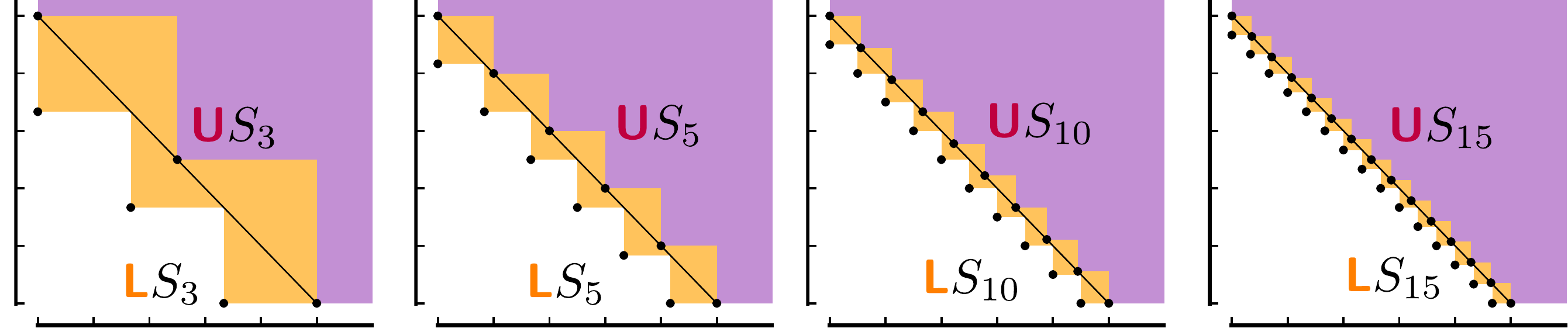}\caption{\label{fig:approx_invplus}Approximations to $\overline{+}$ using
the uniform sampling sequence~$S_{n}$. }
\end{figure}
\par\end{center}

However, the convergence is not monotonic, in the sense that~$S_{n+1}{\not\posleq}_{\udpsp}S_{n}.$
The situation can be represented graphically as in~\prettyref{fig:notchain}.
The sequence~$S_{n}$ eventually converges to $\overline{+}$, but
it is not a descending chain. This means that it is not true, in general,
that the solution to the MCDP obtained by plugging in~$S_{n+1}$
gives smaller bounds than~$S_{n}$.

\subsubsection*{Relaxation based on Van Der Corput sequence}

We can easily create an approximation sequence~$V:\mathbb{N}\rightarrow\udpsp$
that converges monotonically using Var Der Corput (VDC) sampling~\cite[Section 5.2]{LaValle2006Planning}.
Let~$\vdc(n)$ be the VDC sequence of~$n$ elements in the interval~$[0,1]$.
The first elements of the VDC are $\{0,0.5,0.25,0.75,0.125,\dots\}$.
The sequence is guaranteed to satisfy~$\vdc(n)\subseteq\vdc(n+1)$
and to minimize the ``discrepancy'', a measure of uniform coverage.

The upper bound~$\udpU V_{n}$ is defined as sampling the segment
with extrema~$\left\langle 0,\fun_{1}\right\rangle $ and~$\left\langle \fun_{1},0\right\rangle $
using the VDC sequence:
\[
\udpU V_{n}\colon\fun_{1}\mapsto\{\left\langle \fun_{1}x,\fun_{1}(1-x)\right\rangle \mid x\in\vdc(n)\}.
\]
 The lower bound~$\udpL V_{n}$ is defined by taking meets of successive
points, according to the procedure in~\prettyref{fig:make_lower}.
\begin{center}
\begin{figure}[H]
\begin{centering}
\adjustbox{max width=8.6cm}{\includegraphics[scale=0.33]{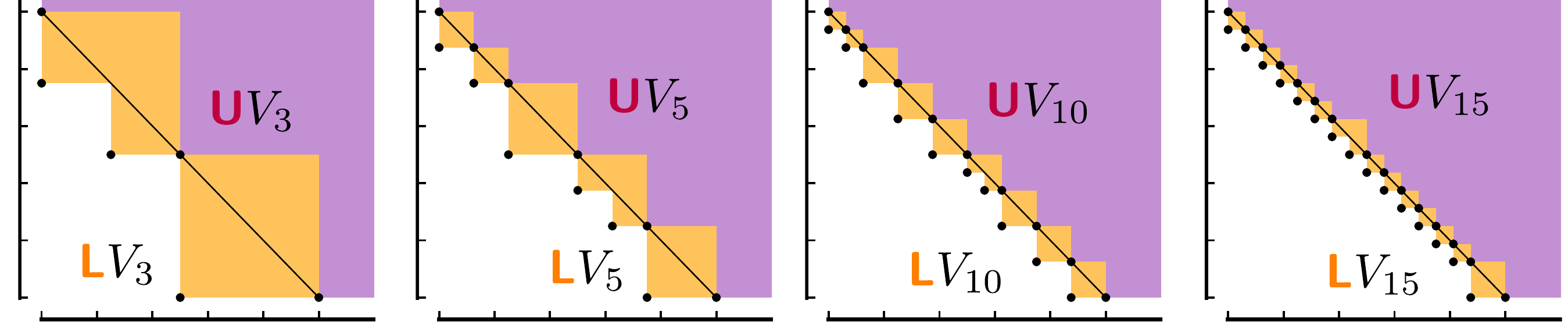}}
\par\end{centering}
\caption{\label{fig:Vn}Approximations to $\overline{+}$ using the Van Der
Corput sequence~$V_{n}$. The orange and purple areas are the upper
closures of the antichains represented by the black points.}
\end{figure}
\par\end{center}

For this sequence, one can prove that not only~$\overline{+}\udpleq V_{n}$,
but also that the convergence is uniform, in the sense that~$\overline{+}\udpleq V_{n+1}\udpleq V_{n}.$
The situation is represented graphically in~\prettyref{fig:convergence_pyramid}:
the sequence is a descending chain that converges to~$\overline{+}$.

\subsubsection{Dual of multiplication}

\begin{figure}[H]
\begin{centering}
\adjustbox{max width=8.6cm}{\includegraphics[scale=0.33]{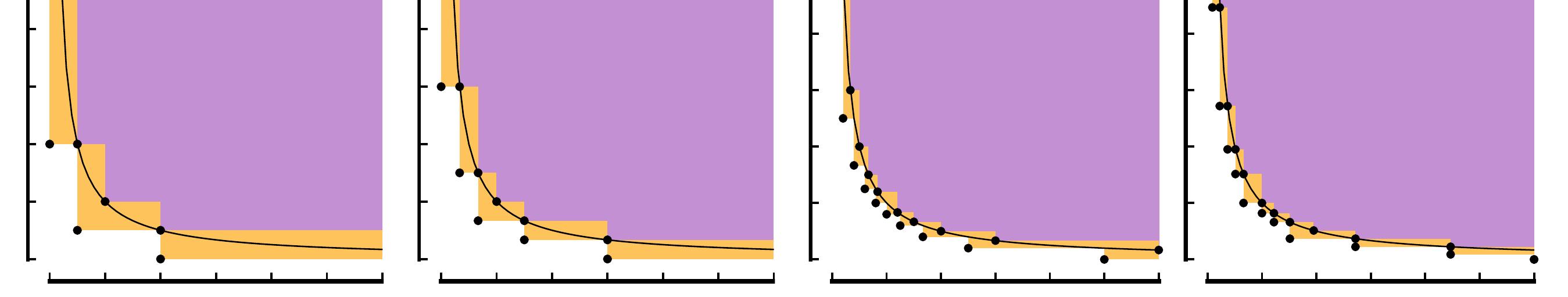}}
\par\end{centering}
\caption{\label{fig:approx_invmult}\C{Van Der Corput relaxations of $\fun_{1}\leq\res_{1}\res_{2}$
for $n=3,5,10,15.$} The orange and purple areas are the upper closures
of the antichains represented by the black points.}
\end{figure}

The case of multiplication can be treated analogously to the case
of addition. By taking the logarithm, the inequality~$\fun_{1}\leq\res_{1}\res_{2}$
can be rewritten as~$\log(\fun_{1})\leq\log(\res_{1})+\log(\res_{2}).$
So we can repeat the constructions done for addition. The VDC sequence
are shown in~\prettyref{fig:approx_invmult}.

\subsubsection{Numerical example}

This relaxation strategy was applied to the relation \emph{${\colF\text{travel distance}}\leq{\colR\text{velocity}}\times{\colR\text{endurance}}$}
in the MCDP in~\prettyref{fig:Example1}. Thanks to the theory, we
can obtain estimates of the solutions using bounded computation, even
though that relation has infinite cardinality.

\prettyref{fig:invplus1}~shows the result using uniform sampling,
and~\prettyref{fig:invplus2} shows the result using VDC sampling.
As predicted by the theory, uniform sampling does not give monotone
convergence, while VDC sampling does.
\begin{center}
\begin{figure}[t]
\begin{centering}
\subfloat[\label{fig:notchain}Qualitative behavior for $S_{n}$]{\centering{}\includegraphics[scale=0.33]{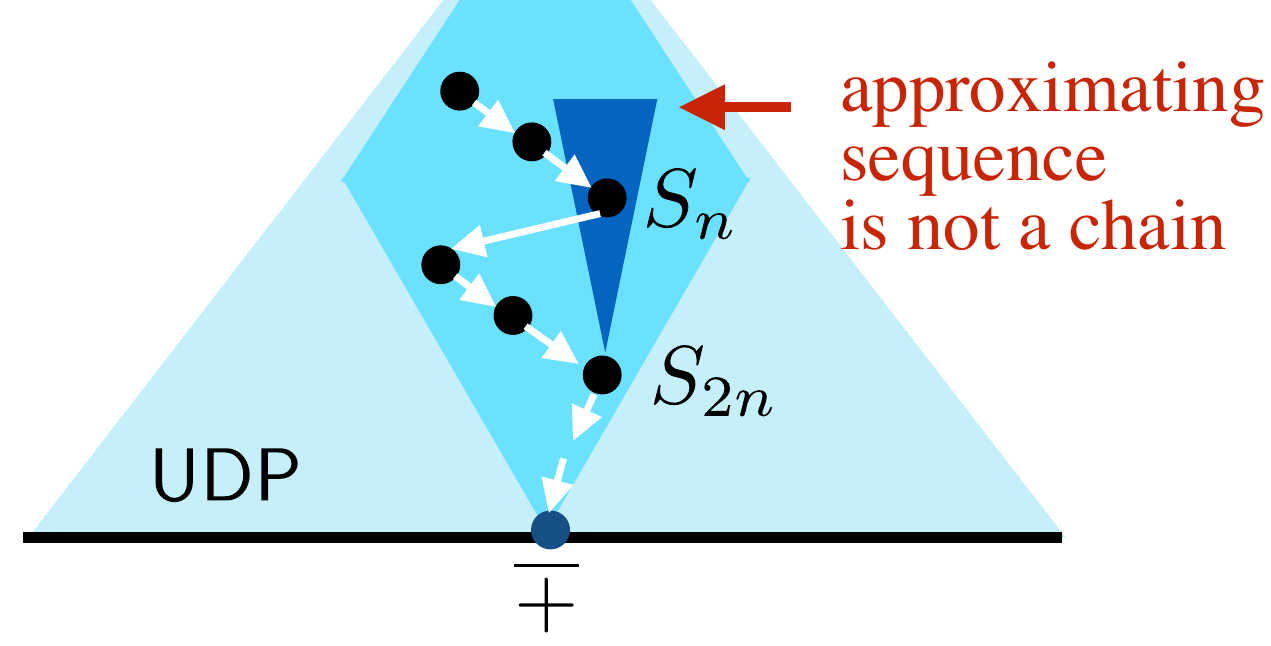}}\subfloat[\label{fig:convergence_pyramid}Qualitative behavior for $V_{n}$]{\begin{centering}
\includegraphics[scale=0.33]{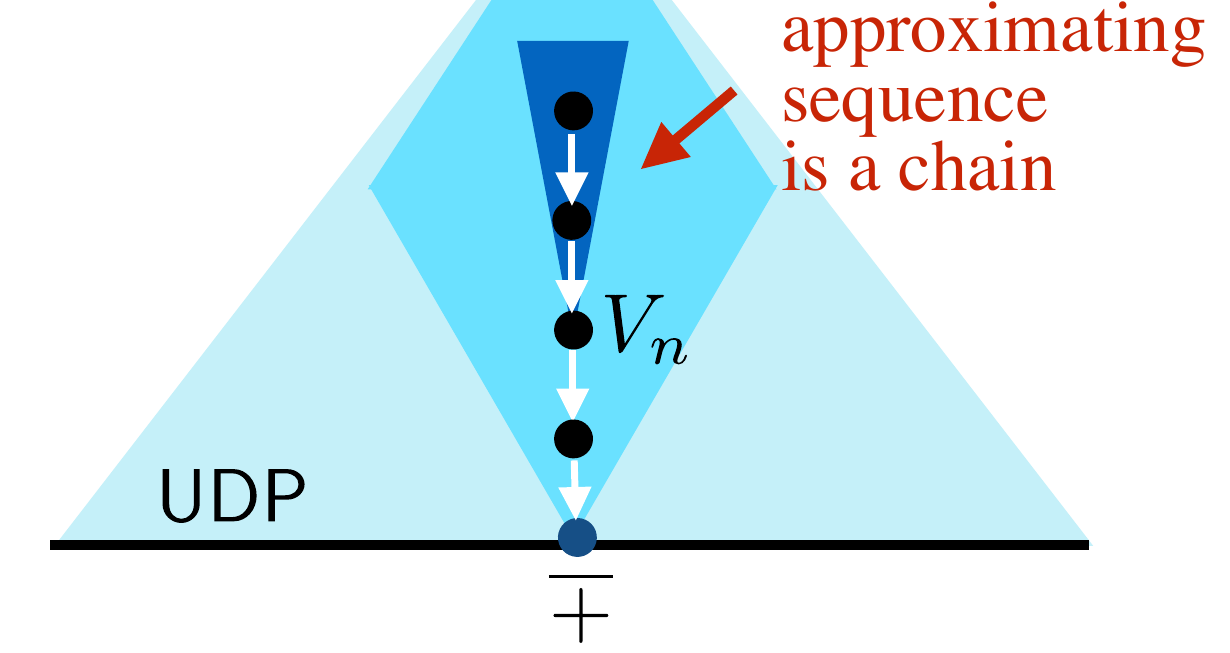}
\par\end{centering}
}
\par\end{centering}
\begin{centering}
\subfloat[\label{fig:invplus1}Numerical results for $S_{n}$]{\begin{centering}
\adjustbox{max width=4.0cm}{\includegraphics[scale=0.33]{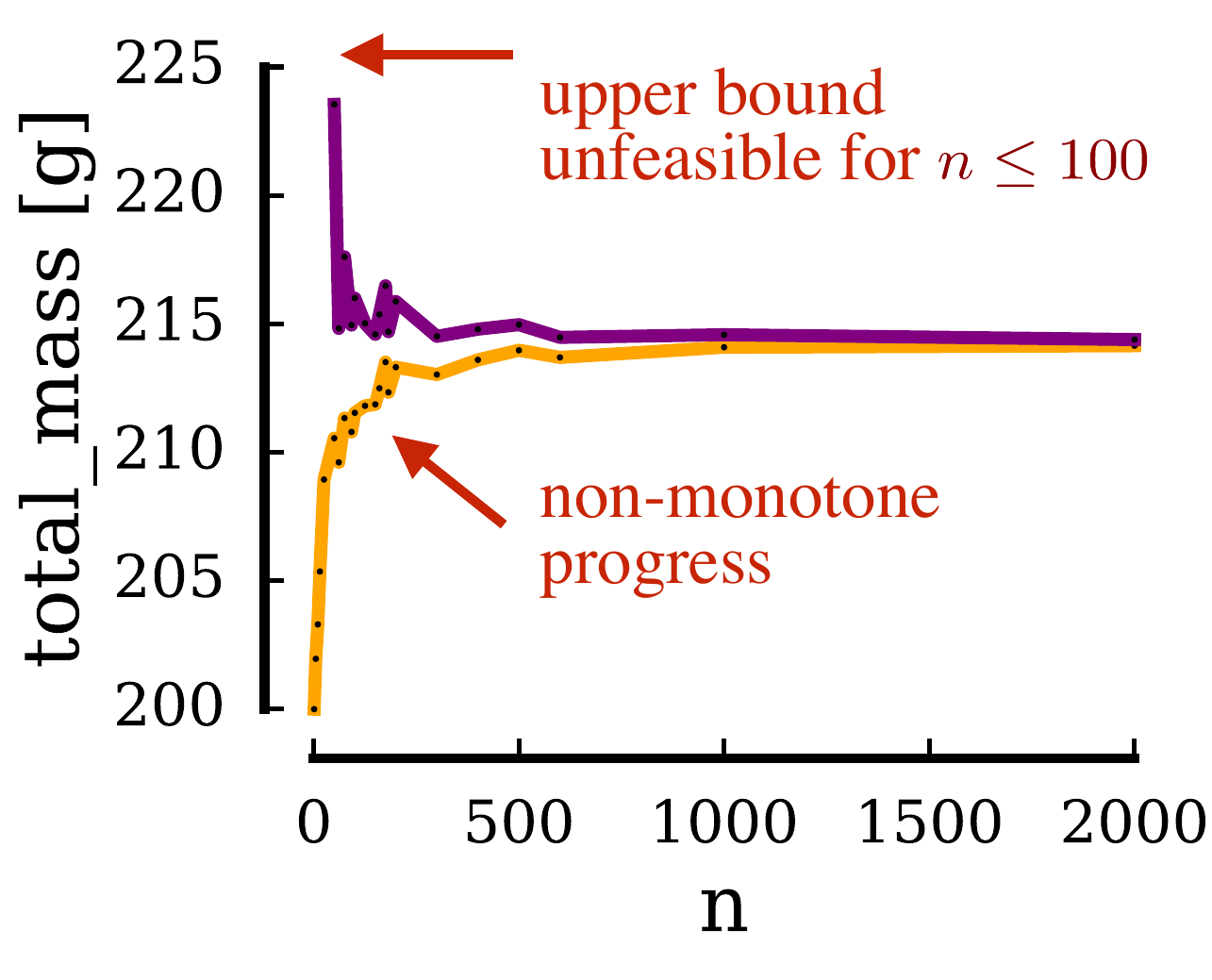}}
\par\end{centering}
}\subfloat[\label{fig:invplus2}Numerical results for $V_{n}$]{\begin{centering}
\adjustbox{max width=4.0cm}{\includegraphics[scale=0.33]{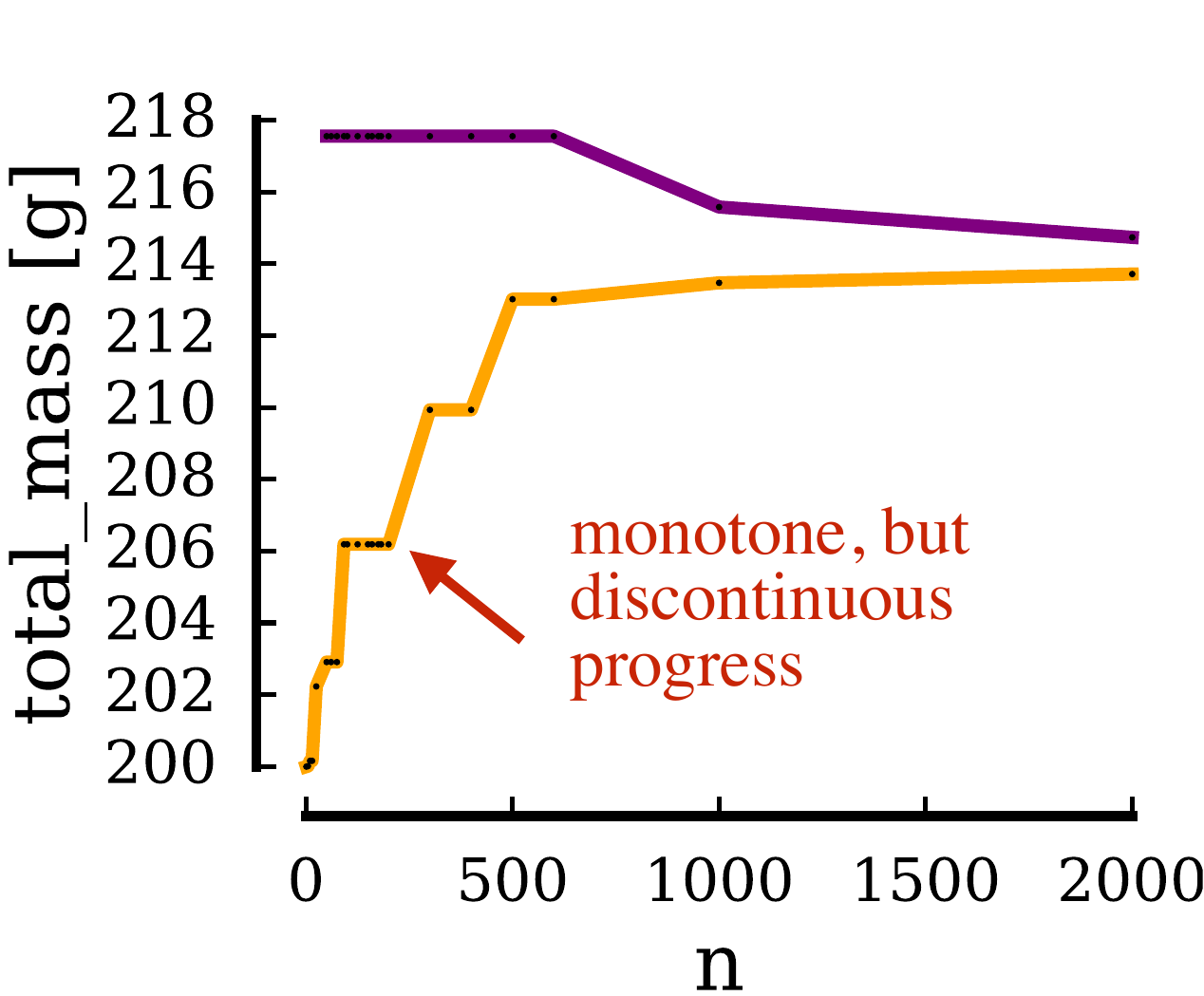}}
\par\end{centering}
}
\par\end{centering}
\caption{Solutions to the example in~\prettyref{fig:Example1}, applying relaxations
for the relation \emph{${\colF\text{travel\_distance}}\leq{\colR\text{velocity}}\times{\colR\text{endurance}}$}
using the uniform sampling sequence and the VDC sampling sequence.
The uniform sampling sequence~$S_{n}$ does not converge monotonically
(panel~\emph{a}); therefore the progress is not monotonic~(panel\emph{~c}).
Conversely, the Van Der Corput sequence~$V_{n}$ is a descending
chain (panel~\emph{b}), which results in monotonic progress (panel~\emph{d}).}
\end{figure}
\par\end{center}

\section{Conclusions and future work}

Monotone Co-Design Problems (MCDPs) provide a compositional theory
of ``co-design'' that describes co-design constraints among different
subsystems in a complex system, such as a robotic system.

This paper dealt with the introduction of uncertainty in the framework,
specifically, interval uncertainty. Uncertainty can be used in two
roles. First, it can be used to describe limited knowledge in the
models. For example, in \prettyref{sec:Application-uncertainty},
we have seen how this can be applied to model mistrust about numbers
from Wikipedia. Second, uncertainty allows to generate relaxations
of the problem. We have seen two applications: introducing an allowed
tolerance in one particular variable (\prettyref{sec:Application-tolerance}),
and dealing with relations with infinite cardinality using bounded
computation resources (\prettyref{sec:Application-relax}).

Future work includes strengthening these results. For example, we
are not able to predict the resulting uncertainty in the solution
before actually computing it; ideally, one would like to know how
much computation is needed (measured by the number of points in the
antichain approximation) for a given value of the uncertainty that
the user can accept.

{

\footnotesize

\setcounter{page}{1}

\printbibliography

}

\appendix

\section{{\normalsize{}Appendix}}

\subsection{Proofs}

\subsubsection{Proofs of well-formedness of \defref{semantics-udp}}

As some preliminary business, we need to prove that \defref{semantics-udp}
is well formed, in the sense that the way the semantics function~$\udpsem$
is defined, it returns a UDP for each argument. This is not obvious
from~\defref{semantics-udp}. 

For example, for $\udpsem\llbracket\atoms,\dpseries(\atree_{1},\atree_{2}),\val\rrbracket$,
the definition gives values for~$\udpL\udpsem\llbracket\atoms,\dpseries(\atree_{1},\atree_{2}),\val\rrbracket$
and~$\udpU\udpsem\llbracket\atoms,\dpseries(\atree_{1},\atree_{2}),\val\rrbracket$
separately, without checking that 
\[
\udpL\udpsem\llbracket\atoms,\dpseries(\atree_{1},\atree_{2}),\val\rrbracket\dpleq\udpU\udpsem\llbracket\atoms,\dpseries(\atree_{1},\atree_{2}),\val\rrbracket.
\]
 The following lemma provides the proof for that.
\begin{lem}
\label{lem:udpsem-well-formed}\defref{semantics-udp} is well formed,
in the sense that {\small{}
\begin{equation}
\udpL\udpsem\llbracket\langle\atoms,\dpseries(\atree_{1},\atree_{2}),\val\rangle\rrbracket\dpleq\udpU\udpsem\llbracket\langle\atoms,\dpseries(\atree_{1},\atree_{2}),\val\rangle\rrbracket,\label{eq:wf1}
\end{equation}
\begin{equation}
\udpL\udpsem\llbracket\left\langle \atoms,\dppar(\atree_{1},\atree_{2}),\val\right\rangle \rrbracket\dpleq\udpU\udpsem\llbracket\left\langle \atoms,\dppar(\atree_{1},\atree_{2}),\val\right\rangle \rrbracket,\label{eq:wf2}
\end{equation}
\begin{equation}
\udpL\udpsem\llbracket\left\langle \atoms,\dploop(\atree),\val\right\rangle \rrbracket\dpleq\udpU\udpsem\llbracket\left\langle \atoms,\dploop(\atree),\val\right\rangle \rrbracket.\label{eq:wf3}
\end{equation}
}{\small \par}
\end{lem}
\begin{IEEEproof}
Proving \eqref{wf1}\textemdash \eqref{wf3} can be reduced to proving
the following three results, for any $x,y\in\udpsp$:
\begin{align*}
\left(\udpL x\opseries\udpL y\right) & \dpleq\left(\udpU x\opseries\udpU y\right),\\
\left(\udpL x\oppar\udpL y\right) & \dpleq\left(\udpU x\oppar\udpU y\right),\\
\left(\udpL x\right)^{\oploop} & \dpleq\left(\udpU x\right)^{\oploop}.
\end{align*}
These are given in \lemref{well-formed-series}, \lemref{well-formed-par},
\lemref{well-formed-loop}.
\end{IEEEproof}
\begin{lem}
\label{lem:well-formed-series}$\left(\udpL x\opseries\udpL y\right)\dpleq\left(\udpU x\opseries\udpU y\right)$. 
\end{lem}
\begin{IEEEproof}
First prove that~$\opseries$ is monotone in each argument (proved
as~\lemref{series-monotone}). Then note that
\[
\left(\udpL x\opseries\udpL y\right)\dpleq\left(\udpL x\opseries\udpU y\right)\dpleq\left(\udpU x\opseries\udpU y\right).
\]
\end{IEEEproof}
\begin{lem}
\label{lem:well-formed-par}$\left(\udpL x\oppar\udpL y\right)\dpleq\left(\udpU x\oppar\udpU y\right)$.
\end{lem}
\begin{IEEEproof}
The proof is entirely equivalent to the proof of~\lemref{well-formed-series}.
First prove that~$\dppar$ is monotone in each argument (proved as~\lemref{par-monotone}).
Then note that~
\[
\left(\udpL x\oppar\udpL y\right)\dpleq\left(\udpL x\oppar\udpU y\right)\dpleq\left(\udpU x\oppar\udpU y\right).
\]
\end{IEEEproof}

\begin{lem}
\label{lem:well-formed-loop}$\left(\udpL x\right)^{\oploop}\dpleq\left(\udpU x\right)^{\oploop}$.
\end{lem}
\begin{IEEEproof}
This follows from the fact that~$\oploop$ is monotone (\lemref{loop-monotone}).
\end{IEEEproof}

\subsubsection{Monotonicity lemmas for DP}

These lemmas are used in the proofs above.
\begin{lem}
\label{lem:series-monotone}$\opseries:\dpsp\times\dpsp\rightarrow\dpsp$
is monotone on~$\langle\dpsp,\dpleq\rangle$.
\end{lem}
\begin{IEEEproof}
In~\defref{opseries}, $\opseries$ is defined as follows for two
maps~$\ftor_{1}\colon\funsp_{1}\rightarrow\Aressp_{1}$ and~$\ftor_{2}\colon\funsp_{2}\rightarrow\Aressp_{2}$:
\[
{\displaystyle \ftor_{1}\opseries\ftor_{2}=\Min_{\posleq_{\ressp_{2}}}\uparrow\bigcup_{s\in\ftor_{1}(\fun)}\ftor_{2}(s)}.
\]
It is useful to decompose this expression as the composition of three
maps:
\[
\ftor_{1}\opseries\ftor_{2}=m\circ g[\ftor_{2}]\circ\ftor_{1},
\]
where~``$\circ$'' is the usual map composition, and~$g$ and~$m$
are defined as follows: 
\begin{align*}
g[\ftor_{2}]:\antichains\ressp_{1} & \rightarrow\upsets\ressp_{2},\\
R & \mapsto\uparrow\bigcup_{s\in R}\ftor_{2}(s),
\end{align*}
and 
\begin{align*}
m:\upsets\ressp_{2} & \rightarrow\antichains\ressp_{2},\\
R & \mapsto\Min_{\posleq_{\ressp_{2}}}R.
\end{align*}

From the following facts:
\begin{itemize}
\item $m$ is monotone.
\item $g[\ftor_{2}]$ is monotone in $\ftor_{2}$.
\item $f_{1}\circ f_{2}$ is monotone in each argument if the other argument
is monotone.
\end{itemize}
Then the thesis follows.
\end{IEEEproof}

\begin{lem}
\label{lem:par-monotone}$\oppar:\dpsp\times\dpsp\rightarrow\dpsp$
is monotone on~$\langle\dpsp,\dpleq\rangle$.
\end{lem}
\begin{IEEEproof}
The definition of $\oppar$ (\defref{opmaps}) is:
\begin{align*}
\ftor_{1}\oppar\ftor_{2}:(\funsp_{1}\times\funsp_{2}) & \rightarrow\antichains(\ressp_{1}\times\ressp_{2}),\\
\left\langle \fun_{1},\fun_{2}\right\rangle  & \mapsto\ftor_{1}(\fun_{1})\times\ftor_{2}(\fun_{2}).
\end{align*}
Because of symmetry, it suffices to prove that $\oppar$ is monotone
in the first argument, leaving the second fixed.

We need to prove that for any two DPs $\ftor_{a},\ftor_{b}$ such
that
\begin{equation}
\ftor_{a}\dpleq\ftor_{b},\label{eq:Ikno}
\end{equation}
and for any fixed $\overline{\ftor}$, then
\[
\ftor_{a}\oppar\overline{\ftor}\dpleq\ftor_{b}\oppar\overline{\ftor}.
\]
Let $R=\overline{\ftor}(\fun_{2})$. Then we have that
\begin{align*}
[\ftor_{a}\oppar\overline{\ftor}](\fun_{1},\fun_{2}) & =\ftor_{a}(\fun_{1})\acprod R,\\{}
[\ftor_{b}\oppar\overline{\ftor}](\fun_{1},\fun_{2}) & =\ftor_{b}(\fun_{1})\acprod R.
\end{align*}
Because of \eqref{Ikno}, we know that 
\[
\ftor_{a}(\fun_{1})\posleq_{\antichains\ressp_{1}}\ftor_{b}(\fun_{1}).
\]
So the thesis follows from proving that the product of antichains
is monotone~(\lemref{product-monotone}).
\end{IEEEproof}
\begin{lem}
\label{lem:product-monotone}The product of antichains~$\acprod:\antichains\ressp_{1}\times\antichains\ressp_{2}\rightarrow\antichains(\ressp_{1}\times\ressp_{2})$
is monotone.
\end{lem}

\begin{lem}
\label{lem:loop-monotone}$\oploop:\dpsp\rightarrow\dpsp$ is monotone
on~$\langle\dpsp,\dpleq\rangle$.
\end{lem}
\begin{IEEEproof}
Let $\ftor_{1}\dpleq\ftor_{2}$. Then we can prove that $\ftor_{1}^{\oploop}\dpleq\ftor_{2}^{\oploop}$.
From the definition of~$\oploop$~(\defref{oploop}), we have that
\begin{align*}
\ftor_{1}^{\oploop}(\fun_{1}) & =\lfp(\Psi_{\fun}^{\ftor_{1}}),\\
\ftor_{2}^{\oploop}(\fun_{2}) & =\lfp(\Psi_{\fun}^{\ftor_{2}}),
\end{align*}
with~$\Psi_{\fun_{1}}^{\ftor}$ defined as 
\begin{align*}
\Psi_{\fun_{1}}^{\ftor}:\Aressp & \rightarrow\Aressp,\\
{\colR R} & \mapsto\Min_{\posleq_{\ressp}}\bigcup_{\res\in{\colR R}}\ftor(\fun_{1},\res)\ \cap\uparrow\res.
\end{align*}
The least fixed point operator $\lfp$ is monotone, so we are left
to check that the map 
\[
\ftor\mapsto\Psi_{\fun_{1}}^{\ftor}
\]
is monotone. That is the case, because if $\ftor_{1}\dpleq\ftor_{2}$
then 
\[
\left[\bigcup_{\res\in{\colR R}}\ftor_{1}(\fun_{1},\res)\ \cap\uparrow\res\right]\posleq_{\Aressp}\left[\bigcup_{\res\in{\colR R}}\ftor_{2}(\fun_{1},\res)\ \cap\uparrow\res\right].
\]
\end{IEEEproof}

\subsubsection{Monotonicity of semantics $\dpsem$}

\begin{lem}[$\dpsem$ is monotone in the valuation]
\label{lem:dpsem-monotone}Suppose that~$\val_{1},\val_{2}:\atoms\rightarrow\dpsp$
are two valuations for which it holds that~$\val_{1}(a)\dpleq\val_{2}(a)$.
Then~$\dpsem\llbracket\left\langle \atoms,\atree,\val_{1}\right\rangle \rrbracket\dpleq\dpsem\llbracket\left\langle \atoms,\atree,\val_{2}\right\rangle \rrbracket$.
\end{lem}
\begin{IEEEproof}
Given the recursive definition of \defref{dpsem}, we need to prove
this just for the base case and for the recursive cases.

The base case, given in (\ref{eq:base}), is
\[
\dpsem\llbracket\left\langle \atoms,a,\val\right\rangle \rrbracket\doteq\val(a),\qquad\text{for all}\ a\in\atoms.
\]
We have
\begin{align*}
\dpsem\llbracket\left\langle \atoms,\atree,\val_{1}\right\rangle \rrbracket & =\val_{1}(a)\\
\dpsem\llbracket\left\langle \atoms,\atree,\val_{2}\right\rangle \rrbracket & =\val_{2}(a)
\end{align*}
and $\val_{1}(a)\dpleq\val_{2}(a)$ by assumption.

For the recursive cases, (\ref{eq:series})\textendash (\ref{eq:loop}),
the thesis follows from the monotonicity of $\opseries$, $\oppar$,
$\oploop$, proved in \lemref{par-monotone}, \lemref{series-monotone},
\lemref{loop-monotone}.
\end{IEEEproof}

\subsubsection{Proof of the main result, \thmref{udpsem-monotone}}

\label{subsec:proof-main-result}

We restate the theorem.

\textbf{Theorem~\ref{thm:udpsem-monotone}}. \emph{If 
\[
\val_{1}\posleq_{V}\val_{2}
\]
then
\[
\udpsem\llbracket\left\langle \atoms,\atree,\val_{1}\right\rangle \rrbracket\udpleq\udpsem\llbracket\left\langle \atoms,\atree,\val_{2}\right\rangle \rrbracket.
\]
}
\begin{IEEEproof}
From the definition of $\udpsem$ and $\dpsem$, we can derive that
\begin{align}
\udpL\udpsem\llbracket\langle\atoms,\atree,\val\rangle\rrbracket & =\dpsem\llbracket\langle\atoms,\atree,\udpL\circ\val\rangle\rrbracket.\label{eq:equiv1}
\end{align}
In particular, for $\val=\val_{1}$,
\begin{equation}
\udpL\udpsem\llbracket\langle\atoms,\atree,\val_{1}\rangle\rrbracket=\dpsem\llbracket\langle\atoms,\atree,\udpL\circ\val_{1}\rangle\rrbracket.\label{eq:w1}
\end{equation}
Because $\val_{1}(a)\udpleq\val_{2}(a),$ from \lemref{dpsem-monotone},
\begin{equation}
\dpsem\llbracket\langle\atoms,\atree,\udpL\circ\val_{1}\rangle\rrbracket\dpleq\dpsem\llbracket\langle\atoms,\atree,\udpL\circ\val_{2}\rangle\rrbracket.\label{eq:w2}
\end{equation}
From~\eqref{equiv1} again, 
\begin{equation}
\dpsem\llbracket\langle\atoms,\atree,\udpL\circ\val_{2}\rangle\rrbracket=\udpL\udpsem\llbracket\langle\atoms,\atree,\val_{2}\rangle\rrbracket.\label{eq:w3}
\end{equation}
From (\ref{eq:w1}),~(\ref{eq:w2}), (\ref{eq:w3}) together, 
\[
\udpL\udpsem\llbracket\langle\atoms,\atree,\val_{1}\rangle\rrbracket\dpleq\udpL\udpsem\llbracket\langle\atoms,\atree,\val_{2}\rangle\rrbracket.
\]
 Repeating the same reasoning for~$\udpU$, we have 
\[
\udpU\udpsem\llbracket\langle\atoms,\atree,\val_{2}\rangle\rrbracket\dpleq\udpU\udpsem\llbracket\langle\atoms,\atree,\val_{1}\rangle\rrbracket.
\]
 Therefore 
\[
\udpsem\llbracket\langle\atoms,\atree,\val_{1}\rangle\rrbracket\udpleq\udpsem\llbracket\langle\atoms,\atree,\val_{2}\rangle\rrbracket.
\]
\end{IEEEproof}

\vfill\pagebreak

\section{Software}

\subsection{Source code}

The implementation is available at the repository \url{http://github.com/AndreaCensi/mcdp/},
in the branch ``uncertainty\_sep16''. 

\subsection{Virtual machine }

A VMWare virtual machine is available to reproduce the experiments
at the URL \url{https://www.dropbox.com/sh/nfpnfgjh9hpcgvh/AACVZfdVXxMoVqTYiHWaOwHAa?dl=0}.

To reproduce the figures, log in with user password ``mcdp''/''mcdp''.
Then execute the following commands:

\footnotesize
\begin{lyxcode}
\$~cd~\textasciitilde{}/mcdp

\$~source~environment.sh

\$~cd~libraries/examples/uav\_energetics/

~~~~~~~droneD\_complete\_templates.mcdplib

\$~make~clean

\$~make~paper-figures
\end{lyxcode}

\clearpage



\section*{Supplementary material}

# C    Supplementary materials: model definition

The figures contained in this Appendix describe a subset of the models used for optimization.

The goal is to give an idea of how a Monotone Co-Design Problem (MCDP) is formalized using the formal language MCDPL.

This Appendix does not explain the syntax of MCDPL. For details, please see the manual available at `http://mcdp.mit.edu`.

## C.1   General template

All the MCDPs used in the experiments are instantiation of the same *template*, whose code is shown in Fig. C.1.

Figure C.1: Template `DroneCompleteTemplate`

```
template [
    Battery: `BatteryInterface,
    Actuation: `ActuationInterface,
    Perception: `PerceptionInterface,
    PowerApprox: `PowerApprox]
mcdp {
  provides travel_distance [km]
  provides num_missions [R]
  provides carry_payload [g]

  requires total_cost_ownership [$]
  requires total_mass [g]

  strategy = instance `droneD_complete_v2.Strategy

  actuation_energetics =
    instance specialize [
      Battery: Battery,
      Actuation: Actuation,
      PowerApprox: PowerApprox
    ] `ActuationEnergeticsTemplate

  actuation_energetics.endurance >= strategy.endurance
  actuation_energetics.velocity >= strategy.velocity
  actuation_energetics.num_missions >= num_missions
  actuation_energetics.extra_payload >= carry_payload
  strategy.distance >= travel_distance

  perception = instance Perception
  perception.velocity >= strategy.velocity

  actuation_energetics.extra_power >= perception.power

  required total_mass >= actuation_energetics.total_mass

  total_cost_ownership >=  actuation_energetics.total_cost
}
```

Note the top-level functionality:

- `travel_distance` [km];
- `num_missions` (unitless);
- `carry_payload` [g]

and the top-level resources:

- `total_mass` [g]
- `total_cost_ownership` [USD]

The template has four parameters:

- `Battery`: MCDP for energetics;
- `Actuation`: MCDP for actuation;
- `Perception`: MCDP for perception;
- `PowerApprox`: MCDP describing the tolerance for the power variable. This is used in Section VII-B.

Every experiment chooses different values for the parameters of this template.

The graphical representation of the template is shown in Fig. C.3. The dotted blue containers represent the "holes" that need to be filled to instantiate the template.

In turn, the template contains a specialization call to another template, called `ActuationEnergeticsTemplate`, whose code is shown in Fig. C.2 and whose graphical representation is shown in Fig. C.4.

Figure C.2: Template `ActuationEnergeticsTemplate`

```
template [
  Battery: `BatteryInterface,
  Actuation: `ActuationInterface,
  PowerApprox: `PowerApprox
] mcdp {
  provides endurance      [s]
  provides extra_payload [kg]
  provides extra_power    [W]
  provides num_missions  [R]
  provides velocity      [m/s]

  requires total_cost [$]

  battery = instance  Battery
  actuation = instance Actuation

  total_power0 = power required by actuation + extra_power

  power_approx = instance PowerApprox
  total_power0 <= power_approx.power
  total_power = power required by power_approx

  capacity provided by battery >= provided endurance * total_power

  total_mass = (
      mass required by battery +
      actuator_mass required by actuation
      + extra_payload)

  gravity = 9.81 m/s^2
  weight = total_mass * gravity

  lift provided by actuation >= weight
  velocity provided by actuation >= velocity

  labor_cost = (10 $) * (maintenance required by battery)

  required total_cost >= (
    cost required by actuation +
    cost required by battery +
    labor_cost)

  battery.missions >= num_missions

  requires total_mass >= total_mass
}
```

The template has functionality `endurance`, `extra_payload`, `extra_power`, `num_missions`, `velocity`, and two resources, `total_mass` and `total_cost`

Figure C.3: Graphical representation for the template `DroneCompleteTemplate` (Fig. **??**)

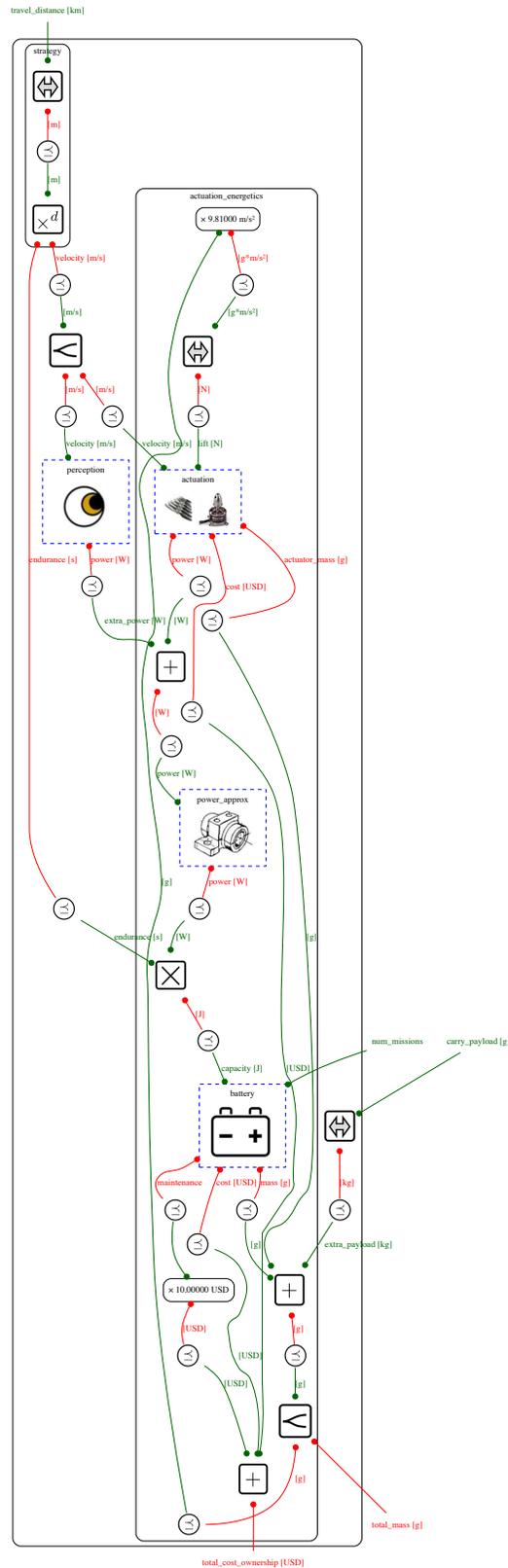

Figure C.4: Graphical representation for the template `ActuationEnergeticsTemplate` (Fig. C.2)

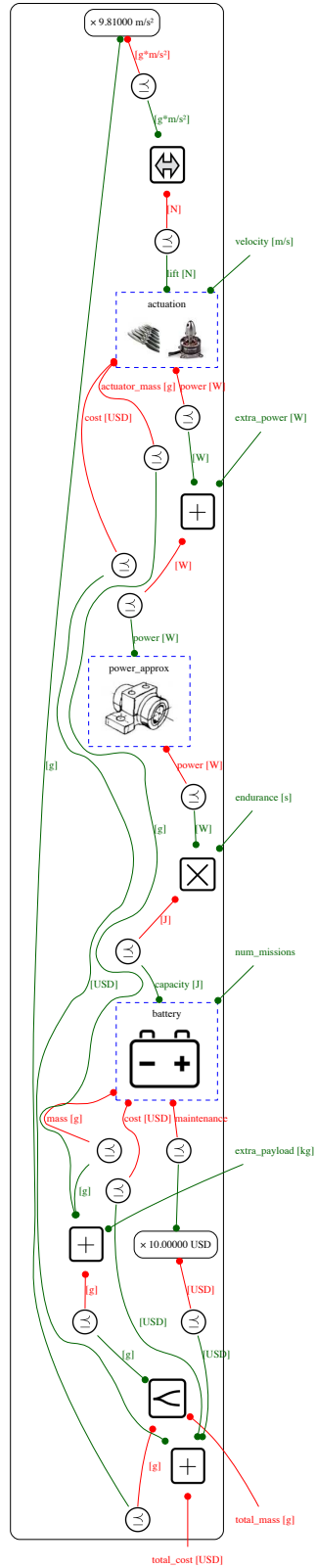

## C.2 MCDP defining batteries properties

Fig. C.5 shows the definition of a single battery technology in terms of specific energy, specific cost, and lifetime (number of cycles).

Figure C.5: Definition of `Battery_LiPo` MCDP

```
 1  mcdp {
 2      provides capacity [J]
 3      provides missions [R]
 4
 5      requires mass      [g]
 6      requires cost      [$]
 7
 8      # Number of replacements
 9      requires maintenance [R]
10
11      # Battery properties
12      specific_energy = 150 Wh/kg
13      specific_cost =   2.50 Wh/$
14      cycles = 600 []
15
16      # Constraint between mass and capacity
17      mass >= capacity / specific_energy
18
19      # How many times should it be replaced?
20      num_replacements = ceil(missions / cycles)
21      maintenance >= num_replacements
22
23      # Cost is proportional to number of replacements
24      cost >= (capacity / specific_cost) * num_replacements
25  }
```

Here a battery is abstracted as a DP with functionality:

- `capacity` [J];
- `missions` (unitless)

and with resources:

- `mass` [g]
- `cost` [USD]

The corresponding graphical representation is shown in Fig. C.6.

Figure C.6: Definition of battery technology

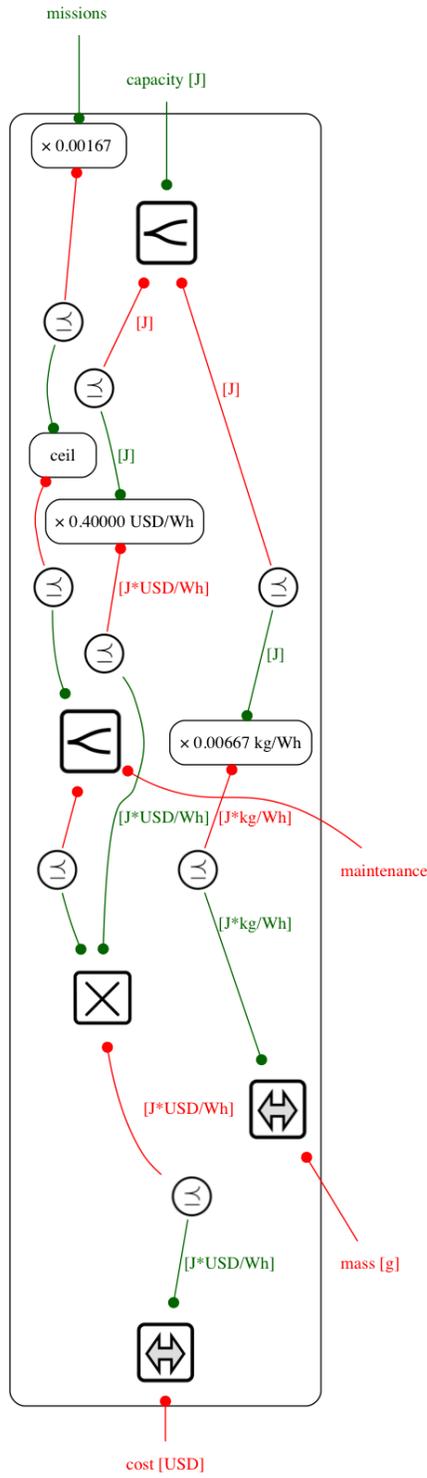

Because this MCDP is completely specified, as opposed to the two *templates* shown earlier, we can show its algebraic representation, as defined in Def. 6.

The MCDP interpreter takes the code shown in Fig. C.5 and then builds an intermediate graphical representation like the one shown in Fig. C.6. Finally, it is compiled to an algebraic representation $\langle \mathcal{A}, \mathbf{T}, v \rangle$, where $\mathbf{T}$ is a tree in the $\{\mathsf{series}, \mathsf{par}, \mathsf{loop}\}$ algebra.

A representation of $\mathbf{T}$ for this example is shown in Fig. C.7.

Each edge is the tree is labeled with the signature of the DP, in the form $\mathcal{F} \to \mathcal{R}$. The junctions are one of the $\{\mathsf{series}, \mathsf{par}, \mathsf{loop}\}$ operators. (The operator $\mathsf{loop}$ does not appear in this example.)

The leaves are labeled with a representation of the Python class that implements them. In particular, the frequently-appearing `Mux` type represents various multiplexing operations, such as

$$\langle x, y, z \rangle \mapsto \langle \langle z, y \rangle, x \rangle.$$

These are necessary to transform a graph into a tree representation.

Figure C.7: Algebraic representation for the example in Fig. C.5

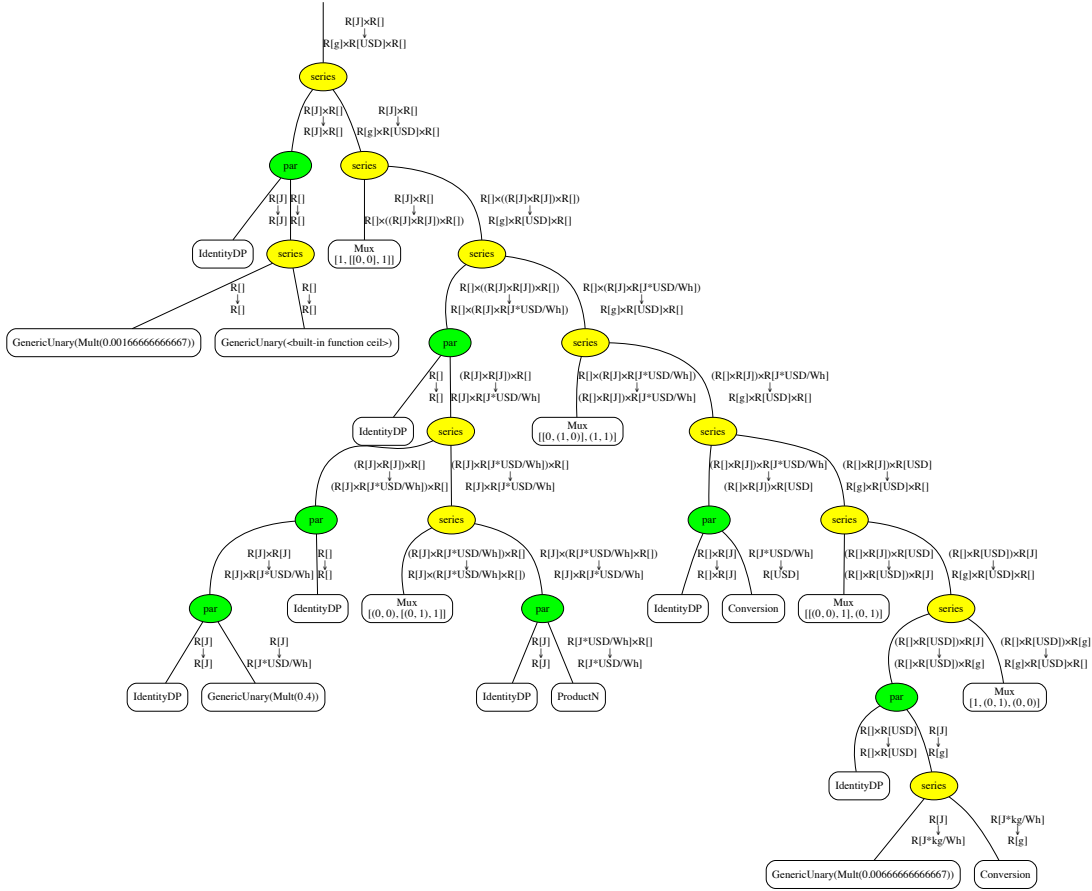

## C.3 Choice between different batteries

Just like we defined `Battery_LiPo` (Fig. C.5), other batteries technologies are similarly defined, such as `Battery_NiMH`, `Battery_LCO`, etc.

Then we can easily express the choice between any of them using the keyword **choose**, as in Fig. C.8.

Figure C.8: Definition of the `Batteries` MCDP

```
1  choose(
2          NiMH: (load Battery_NiMH),
3          NiH2: (load Battery_NiH2),
4           LCO: (load Battery_LCO),
5           LMO: (load Battery_LMO),
6          NiCad: (load Battery_NiCad),
7           SLA: (load Battery_SLA),
8          LiPo: (load Battery_LiPo),
9           LFP: (load Battery_LFP)
10 )
```

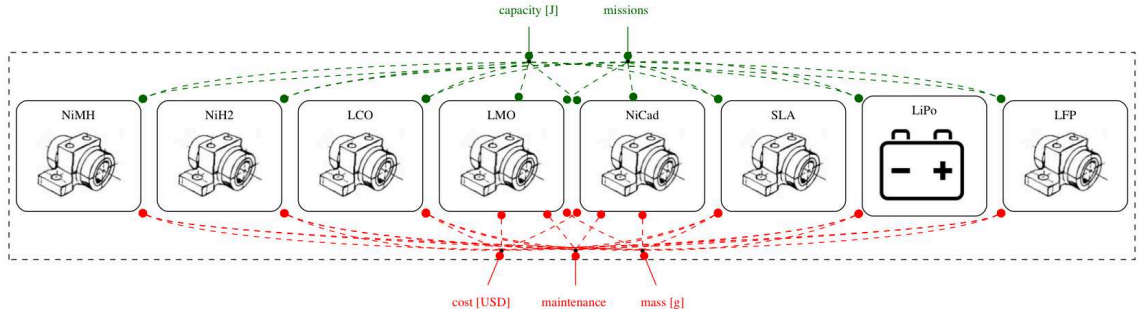

The choice between different batteries is modeled by a *coproduct* operator. This is another type of junction, in addition to series, par, loop that was not described in the paper.

Formally, the coproduct operator it is defined as follows:

$$h_1 \sqcup \cdots \sqcup h_n : \mathcal{F} \to \mathsf{A}\mathcal{R}, \tag{1}$$

$$\mathsf{f} \mapsto \underset{\preceq_\mathcal{R}}{\mathrm{Min}} \left( h_1(\mathsf{f}) \cup \cdots \cup h_n(\mathsf{f}) \right). \tag{2}$$

The algebraic representation (Fig. C.9) contains then one branch for each type of battery.

Figure C.9: Algebraic representation of the `Batteries` MCDP

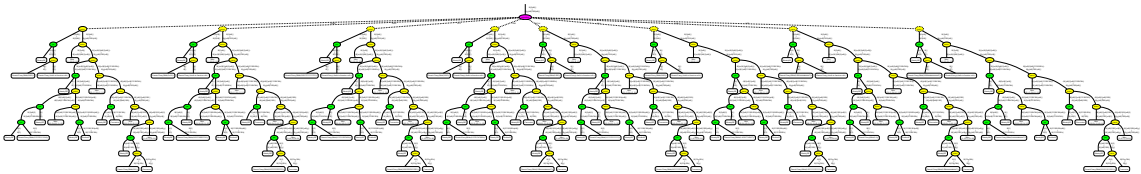

## C.4   Describing uncertainty

This is a description of the Uncertain MCDPs used in the experiments in Section VII-A.

MCDPL has an `Uncertain` operator that can describe interval uncertainty.

For example, the MCDP in Fig. C.5 is rewritten with uncertainty to obtain the code in Fig. C.10. The figures have a 5% uncertainty added to them.

Figure C.10: Definition of `Battery_LiPo` MCDP with 5% uncertainty on parameters

```
 1  mcdp {
 2      provides capacity [J]
 3      provides missions [R]
 4
 5      requires mass      [g]
 6      requires cost      [USD]
 7
 8      # Number of replacements
 9      requires maintenance [R]
10
11      # Battery properties
12      specific_energy_inv = Uncertain(1.0 []/ 157.5 Wh/kg, 1.0 [] /  142.5 Wh/kg)
13      specific_cost_inv = Uncertain(1.0 [] / 2.625 Wh/USD, 1.0 [] / 2.375 Wh/USD)
14      cycles_inv = Uncertain(1.0 []/630.0 [], 1.0[]/ 570.0 [])
15
16      # Constraint between mass and capacity
17      massc = provided capacity * specific_energy_inv
18
19      # How many times should it be replaced?
20      num_replacements = ceil(provided missions * cycles_inv)
21      required maintenance >= num_replacements
22
23      # Cost is proportional to number of replacements
24      costc = (provided capacity * specific_cost_inv) * num_replacements
25
26      required cost >= costc
27      required mass >= massc
28  }
```

In the graphical representation, the uncertainty is represented as "uncertainty gates" that have two branches: one for best case and one for worst case (Fig. C.11).

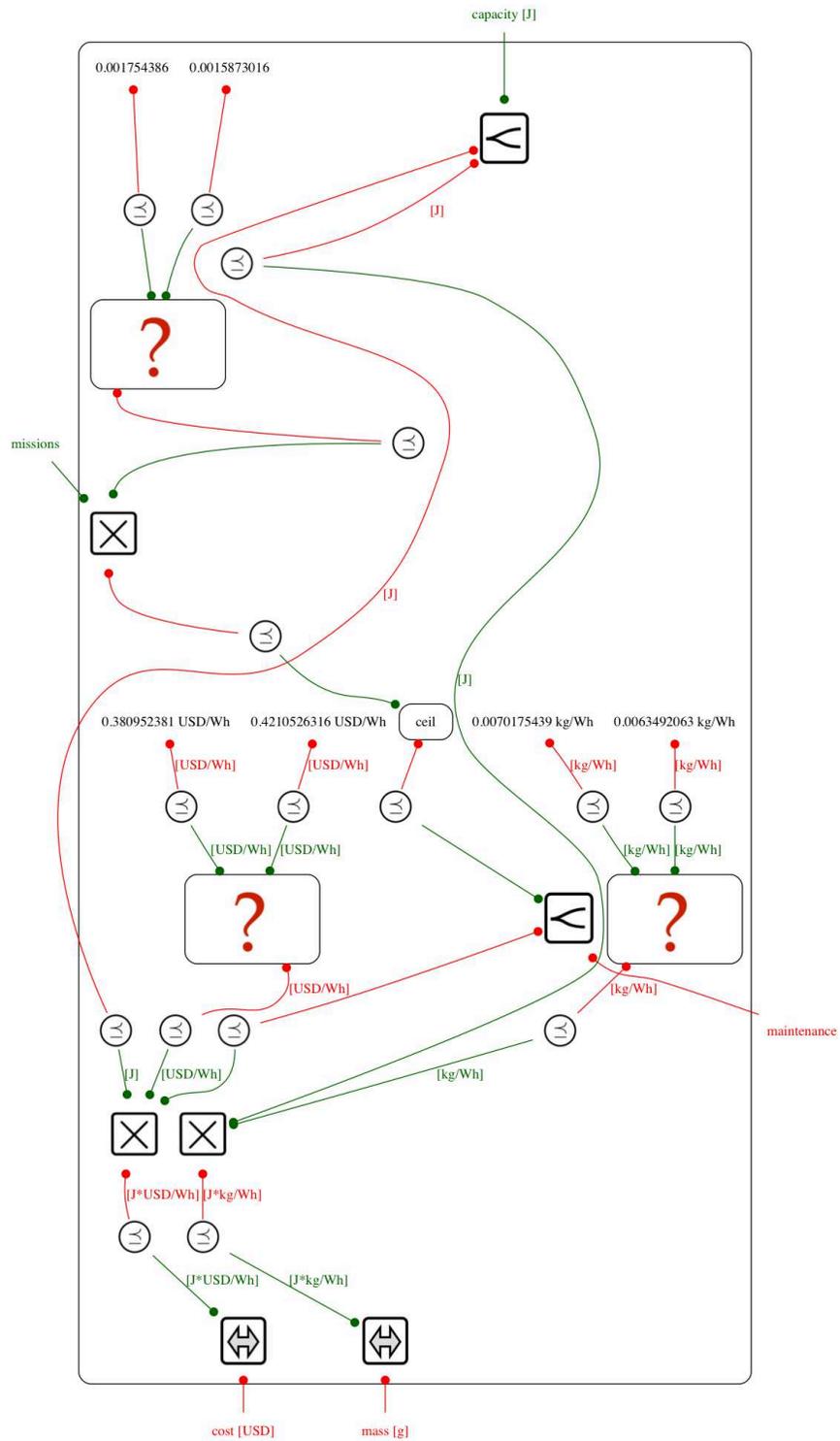

Figure C.11: Graphical representation of uncertain MCDP in Fig. C.10

## C.5 Specialization of templates

Once all the single pieces are defined, then the final MCDP is assembled using the `specialize` keyword.

For example, the following code specializes the template using only the `Battery_LiPo` MCDP.

Figure C.12:

```
specialize [
    Battery: `batteries_nodisc.Battery_LiPo,
    Actuation: `droneD_complete_v2.Actuation,
    Perception: `Perception1,
    PowerApprox: `PowerApprox
]
`DroneCompleteTemplate
```

The algebraic representation is shown in Fig. C.13.

Figure C.13: Algebraic representation of MCDP in Fig. C.12

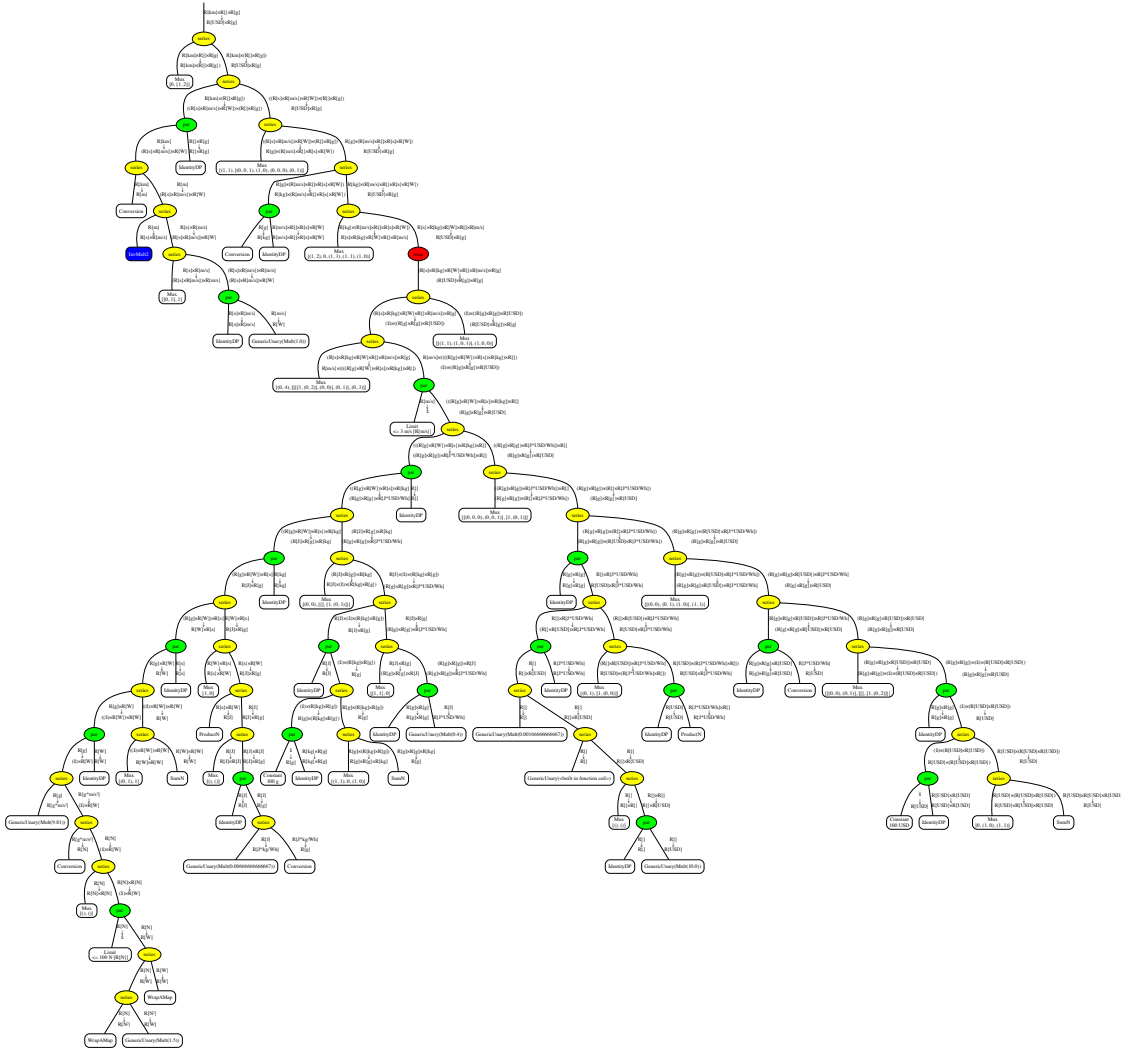

The following code specializes the template using the coproduct of all batteries, each having an uncertain specification.

Figure C.14:

```
 1  specialize [
 2    Battery: `batteries_uncertain1.batteries,
 3    Actuation: `droneD_complete_v2.Actuation,
 4    PowerApprox: mcdp {
 5      provides power [W]
 6      requires power [W]
 7
 8      required power >= approxu(provided power, 1 mW)
 9    }
10  ] `ActuationEnergeticsTemplate
```

The algebraic representation is shown in Fig. C.15.

The blue nodes are the uncertain nodes (UDPs).

Figure C.15: Algebraic representation of MCDP in Fig. C.14

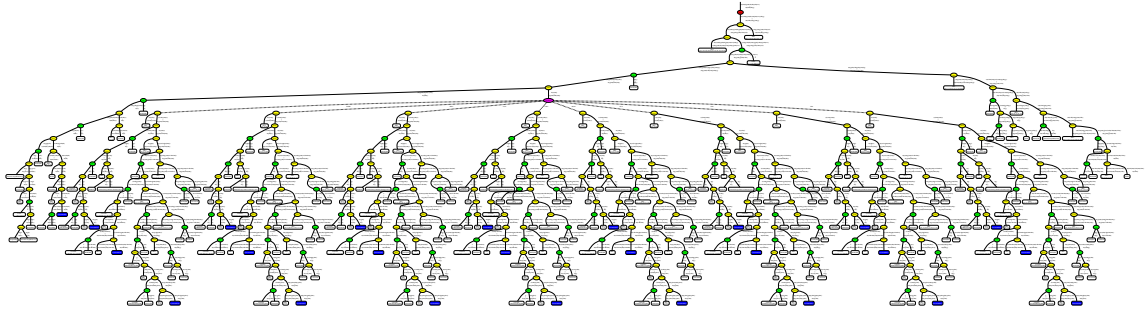

 

\end{document}